%% file: cvpr.tex
\documentclass[10pt,twocolumn,letterpaper]{article}

 \usepackage[pagenumbers]{cvpr} 

\usepackage{graphicx}
\usepackage{amsmath}
\usepackage{amssymb}
\usepackage{booktabs}
\usepackage{multirow}

\usepackage{bm}
\usepackage{nicefrac}

\let\originalleft\left
\let\originalright\right
\def\left#1{\mathopen{}\originalleft#1}
\def\right#1{\originalright#1\mathclose{}}

\makeatletter
\newcommand{\smallsym}[2]{#1{\mathpalette\make@small@sym{#2}}}
\newcommand{\make@small@sym}[2]{%
  \vcenter{\hbox{$\m@th\downgrade@style#1#2$}}%
}
\newcommand{\downgrade@style}[1]{%
  \ifx#1\displaystyle\scriptstyle\else
    \ifx#1\textstyle\scriptstyle\else
      \scriptscriptstyle
  \fi\fi
}

\newcommand{\ucomma}{%
  \leavevmode 
  \kern-0.08em\text{,}\kern0.08em 
}

\makeatother

\usepackage[pagebackref,breaklinks,colorlinks]{hyperref}

\usepackage[capitalize]{cleveref}
\crefname{section}{Sec.}{Secs.}
\Crefname{section}{Section}{Sections}
\Crefname{table}{Table}{Tables}
\crefname{table}{Tab.}{Tabs.}

\def\cvprPaperID{3738} 
\def\confName{CVPR}
\def\confYear{2024}

\begin{document}

\title{Neural Spline Fields for Burst Image Fusion and Layer Separation}


\author{Ilya Chugunov \hspace{1em} David Shustin \hspace{1em} Ruyu Yan \hspace{1em} Chenyang Lei \hspace{1em} Felix Heide \vspace{0.5em}\\
Princeton University}

\maketitle

\begin{abstract}
Each photo in an image burst can be considered a sample of a complex 3D scene: the product of parallax, diffuse and specular materials, scene motion, and illuminant variation. While decomposing all of these effects from a stack of misaligned images is a highly ill-conditioned task, the conventional align-and-merge burst pipeline takes the other extreme: blending them into a single image. In this work, we propose a versatile intermediate representation: a two-layer alpha-composited image plus flow model constructed with neural spline fields -- networks trained to map input coordinates to spline control points. Our method is able to, during test-time optimization, jointly fuse a burst image capture into one high-resolution reconstruction and decompose it into transmission and obstruction layers. Then, by discarding the obstruction layer, we can perform a range of tasks including seeing through occlusions, reflection suppression, and shadow removal. Validated on complex synthetic and in-the-wild captures we find that, with no post-processing steps or learned priors, our generalizable model is able to outperform existing dedicated single-image and multi-view obstruction removal approaches. 

\end{abstract}

\input{0_introduction}
\input{1_related_work}
\input{2_method}
\input{3_results}
\input{4_conclusion}

{\small
\bibliographystyle{ieee_fullname}
\bibliography{cvpr}
}

\input{supp_0_implementation}

\input{supp_1_results}
\input{supp_2_experiments}


\end{document}

%% file: 0_introduction.tex
 \vspace{-1em}
\section{Introduction}
%
Over the last decade, as digital photos have increasingly been produced by smartphones, smartphone photos have increasingly been produced by burst fusion. To compensate for less-than-ideal camera hardware -- typically restricted to a footprint of less than 1cm$^3$~\cite{blahnik2021smartphone} -- smartphones rely on their advanced compute hardware to process and fuse multiple lower-quality images into a high-fidelity photo~\cite{delbracio2021mobile}. This proves particularly important in low-light and high-dynamic-range settings~\cite{liba2019handheld, hasinoff2016burst}, where a single image must compromise between noise and motion blur, but multiple images afford the opportunity to minimize both~\cite{kalantari2017deep}. But even as mobile night- and astro-photography applications~\cite{google2018night,google2019astrophotography} use increasingly long sequences of photos as input, their output remains a static single-plane image. Given the typically non-static and non-planar nature of the real world, a core problem in burst image pipelines is thus the alignment~\cite{lecouat2022high, mildenhall2018burst} and aggregation~\cite{wronski2019handheld, bhat2021deep} of pixels into an image array -- referred to as the \textit{align-and-merge} process.

\begin{figure}[t]
    \centering
    \includegraphics[width=\linewidth]{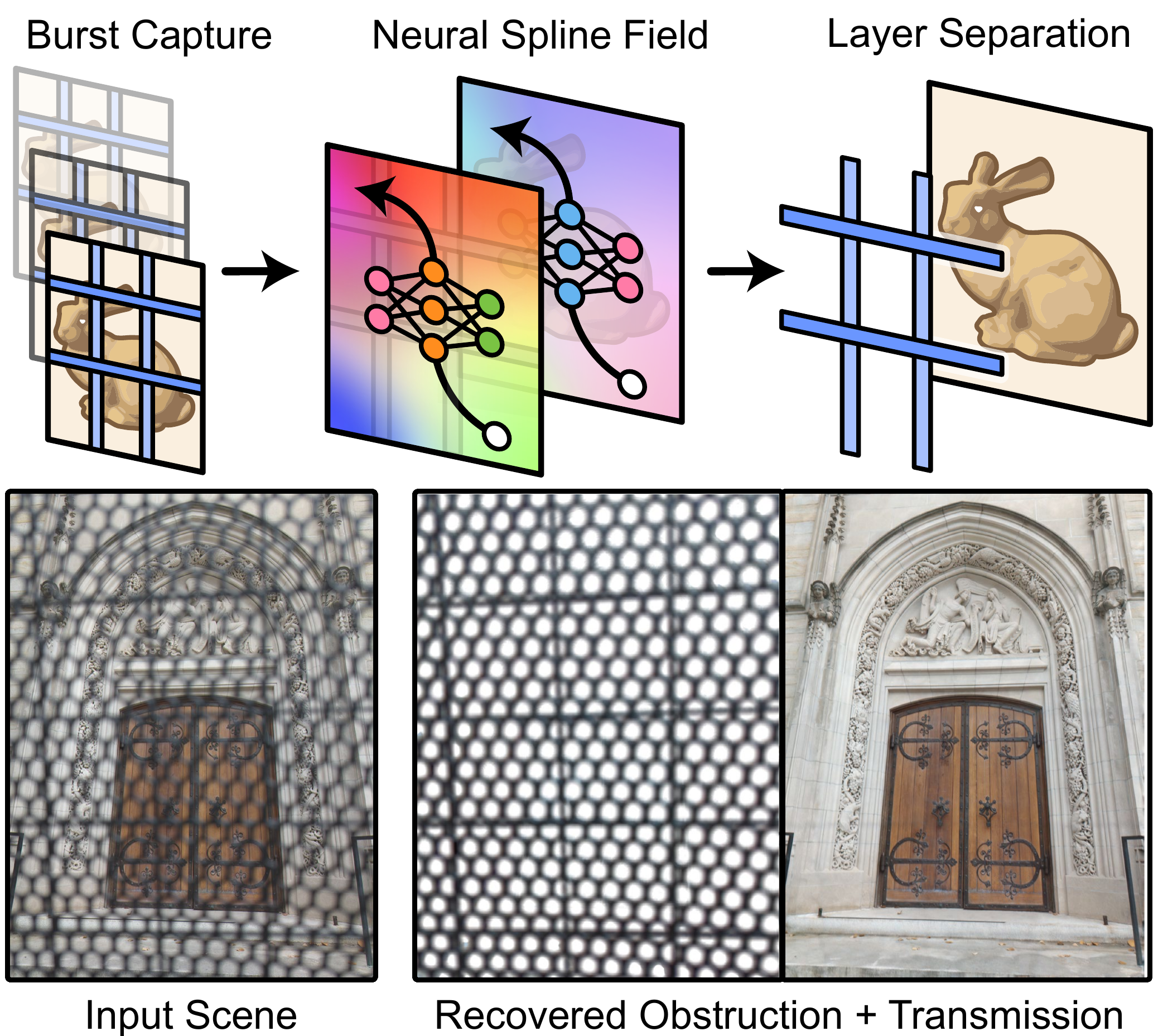}
    \vspace*{-1.5em}
    \caption{Fitting our two-layer neural spline field model to a stack of images we're able to directly estimate and separate even severe, out-of-focus obstructions to recover hidden scene content.}
    \label{fig:teaser}
    \vspace{-1.5em}
\end{figure}

While existing approaches treat pixel motion as a source of noise and artifacts, a parallel direction of work~\cite{chugunov2022implicit,yu20143d,ha2016high} attempts to extract useful parallax cues from this pixel motion to estimate the geometry of the scene. Recent work by Chugunov et al.~\cite{chugunov2023shakes} finds that maximizing the photometric consistency of an RGB plus depth neural field model of an image sequence is enough to distill dense depth estimates of the scene. While this method is able to jointly estimate high-quality camera motion parameters, it does not perform high-quality image reconstruction, and rather treats its image model as ``a vehicle for depth optimization"~\cite{chugunov2023shakes}. In contrast, work by Nam et al.~\cite{nam2022neural} proposes a neural field fitting approach for multi-image fusion and layer separation which focuses on the quality of the reconstructed ``canonical view". By swapping in different motion models, they can separate and remove layers such as occlusions, reflections, and moir\'e patterns during image reconstruction -- as opposed to in a separate post-processing step~\cite{shih2015ghosting,gupta2019fully}. This approach, however, does not make use of a realistic camera projection model, and relies on regularization penalties to discourage its motion models from representing non-physical effects -- e.g., pixel tearing or teleportation.

In this work, we propose a versatile layered neural image representation~\cite{nam2022neural} with a projective camera model~\cite{chugunov2023shakes} and novel neural spline flow parametrization. Our model takes as input an unstabilized 12-megapixel RAW image sequence, camera metadata, and gyroscope measurements -- available on all modern smartphones. During test-time optimization, it fits to produce a high-resolution reconstruction of the scene, separated into \textit{transmission} and \textit{obstruction} image planes. The latter of which can be removed to perform occlusion removal, reflection suppression, or shadow removal. To this end, we decompose pixel motion between burst frames into planar motion, from the camera's pose change in 3D space relative to the image planes, and generic flow components which account for depth parallax, scene motion, and other effects. We model these flows with neural spline fields (NSFs): networks trained to map input coordinates to spline control points, which are then interpolated at sample timestamps to produce flow field values. As their output dynamics are strictly bound by their spline parametrization, these NSFs produce \textit{temporally} consistent flow with no regularization, and can be controlled \textit{spatially} through the manipulation of their positional encodings.

\noindent In summary, we make the following contributions: \vspace{-0.5em}
\begin{itemize}
    \item  An end-to-end neural scene fitting approach which fits to a burst image sequence to distill high-fidelity camera poses, and high-resolution two layer transmission plus occlusion image decomposition. \vspace{-0.5em}
    \item A compact, controllable neural spline field model to estimate and aggregate pixel motion between frames. \vspace{-0.5em}
    \item Qualitative and quantitative evaluations which demonstrate that our model outperforms existing single image and multi-frame obstruction removal approaches. \vspace{-0.5em}
\end{itemize}
Code, data, videos, and additional materials are available on our project website: \href{https://light.princeton.edu/publication/nsf/}{light.princeton.edu/nsf}

%% file: 1_related_work.tex
\section{Related Work} 

\noindent\textbf{Burst Photography.}\hspace{0.1em}
A large body of work has explored methods for burst image processing~\cite{delbracio2021mobile} to achieve high image quality in mobile photography settings. During burst imaging, the device records a sequence of frames in rapid succession -- potentially a \textit{bracketed sequence} with varying exposure parameters~\cite{mertens2009exposure} -- and fuses them post-capture to produce a demosaiced~\cite{tan2017joint}, denoised~\cite{mildenhall2018burst, godard2018deep}, superresolved~\cite{wronski2019handheld,lecouat2022high}, or otherwise enhanced reconstruction. Almost all modern smartphone devices rely on burst photography for low-light~\cite{liba2019handheld,hasinoff2016burst} and high dynamic range reconstruction from low dynamic range sensors~\cite{hasinoff2016burst,gallo2015locally}. While existing methods typically use sequences of only 2-8 frames as input, a parallel field of micro-video~\cite{yu20143d,im2015high} or ``long-burst photography"~\cite{chugunov2023shakes} research -- which also encompasses widely deployed Apple Live Photos, Android Motion Photos, and night photography~\cite{google2018night,google2019astrophotography} -- consumes sequences of images up to several seconds in length, acquired naturally during camera viewfinding. Though not limited to long-burst photography, we adopt this setting to leverage the parallax~\cite{xue2015computational} and pixel motion cues in these extended captures for separation of obstructed and transmitted scene content.

\noindent\textbf{Obstruction Removal and Layer Separation}\hspace{0.1em} 
While their use of visual cues is diverse -- e.g., identifying reflections from ``ghosting" cues on thick glass~\cite{shih2015ghosting} or detecting lattices for fence deletion~\cite{park2011image} -- all single-image obstruction removal is fundamentally a segmentation~\cite{liu2022semantic, kume2023singlefft} and image recovery~\cite{hu2023single, gandelsman2019double} problem. In the most severe cases, with fully opaque occluders, this image recovery problem becomes an in-painting task~\cite{xiong2019foreground, farid2016image} to synthesize missing content. This is in contrast to approaches which rely on multiple measurements such as multi-focal stacks~\cite{shen2023light,adeel2022defencing}, multi-view images~\cite{niklaus2020learned,liu2020learning}, flash no-flash pairs~\cite{lei2021robust,lei2022robustwild}, or polarization data~\cite{lei2020polarized}. These methods typically treat obstruction removal as an inverse problem~\cite{bertero2021introduction}, estimating a model of transmitted and occluded content consistent with observed data~\cite{li2013exploiting}. This can also be generalized to an image layer separation problem, an example of which is intrinsic decomposition~\cite{chen2013simple}, where the separated layer is the obstruction. These methods typically rely on learned priors~\cite{gandelsman2019double} and pixel motion~\cite{nam2022neural} to decompose images into multiple components. Our work explores the layer separation problem in the burst photography setting, where pixel motion is on a much smaller scale than in video sequences~\cite{ye2022deformable}, and a high-resolution unobstructed view is desired as an output. Rather than tailor to a single application, however, we propose a unified model with applications to reflection, occlusion, and shadow separation.

\noindent\textbf{Neural Scene Representations.}\hspace{0.1em} A growing body of work investigating novel view synthesis has demonstrated that coordinate-based neural representations are capable of reconstructing complex scenes~\cite{barron2023zip,barron2021mip} without an explicit structural backbone such as a pixel array or voxel grid. These networks are typically trained from scratch, through \textit{test-time optimization}, on a single scene to map input coordinate encodings~\cite{tancik2020fourier} to outputs such as RGB~\cite{sitzmann2020implicit}, depth~\cite{chugunov2022implicit}, or x-ray data~\cite{sun2021coil}. While neural scene representations require many network evaluations to generate outputs, as opposed to explicit representations which can be considered ``pre-evaluated", recent works have shown great success in accelerating training~\cite{muller2022instant} and inference~\cite{yu2021plenoctrees} of these networks. Furthermore, this per-output network evaluation is what lends to their versatility, as they can be optimized through auto-differentiation with no computational penalties for sparse or non-uniform sampling of the scene~\cite{kjolstad2017tensor}. Several recent approaches make use of neural scene representations in tandem with continuous motion estimation models to fit multi-image~\cite{chugunov2023shakes} and video~\cite{li2023dynibar} data, potentially decomposing it into multiple layers in the process~\cite{nam2022neural, kasten2021layered}. Our work proposes a novel neural spline field continuous flow representation with a projective camera model to separate effects such as occlusions, reflections, and shadows. In contrast to existing approaches, our flow model does not require regularization to prevent overfitting, as its representation power is controlled directly through encoding and spline hyperparameters.

%% file: 2_method.tex
\section{Neural Spline Fields for Burst Photography}

\noindent We begin with a discussion of the proposed neural spline field model of optical flow. We then continue with our full two-layer projective model of burst photography, its loss functions, training procedure, and data collection pipeline.
\subsection{Neural Spline Fields.}
\begin{figure}[t]
    \centering
    \includegraphics[width=\linewidth]{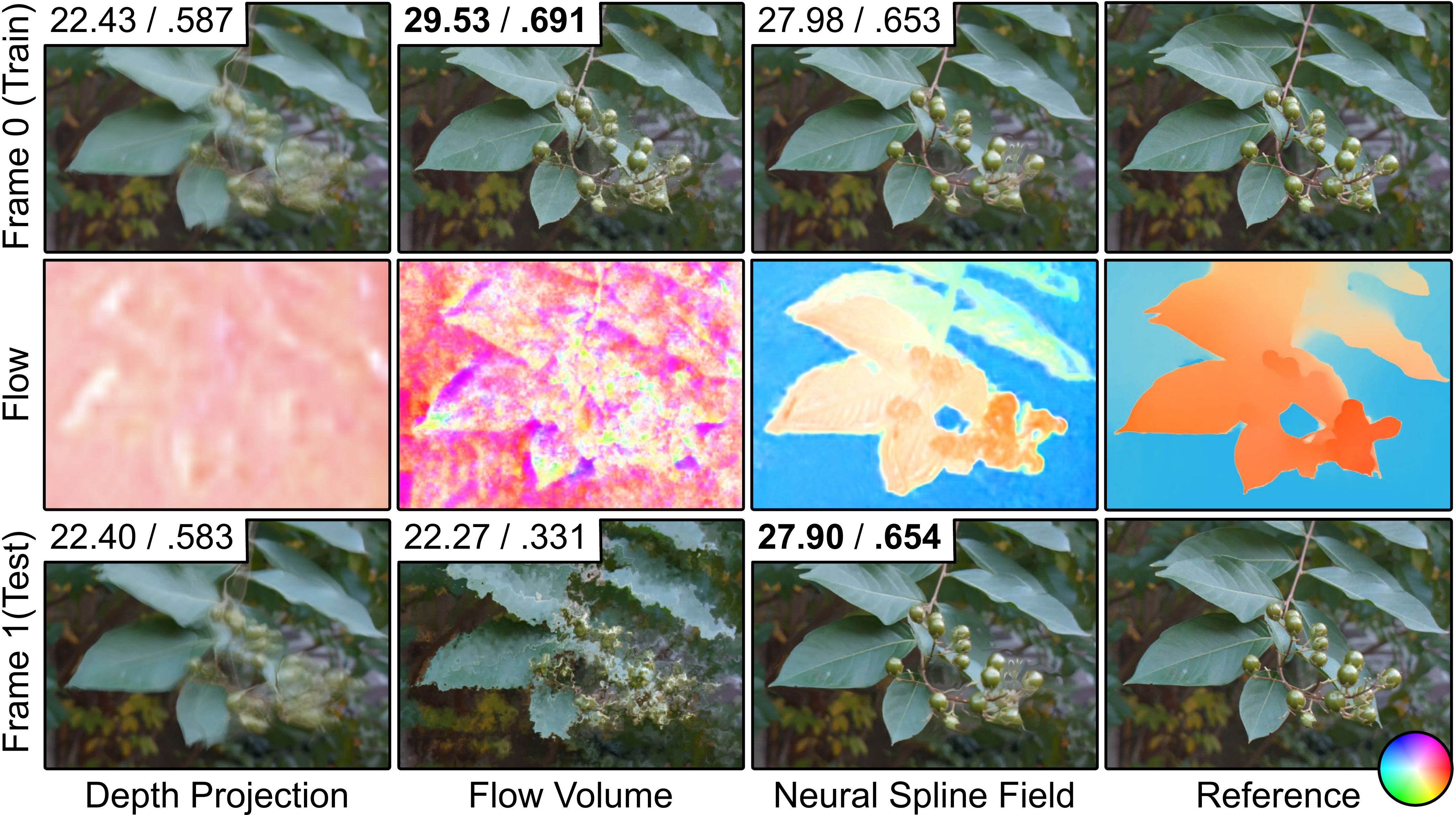}
    \vspace{-1.75em}
    \caption{Image and flow estimates for different representations of a short video sequence of a swinging branch; PSNR/SSIM values inset top-left. Depth projection alone is unable to represent both parallax and scene motion, mixing reconstructed content, and an un-regularized 3D flow volume $g(u,v,t)$ trivially overfits to the sequence. With an identical network, spatial encoding, loss function, and training procedure as $g(u,v,t)$, our neural spline field $S(t; \mathbf{P} = h(u,v))$ produces temporally consistent flow estimates well-correlated with a conventional optical flow reference~\cite{lipson2021raft}.}
    \label{fig:flow_representations}
    \vspace{-1em}
\end{figure}

\noindent\textbf{Motivation.}\hspace{0.1em} To recover a latent image, existing burst photography methods \textit{align and merge}~\cite{delbracio2021mobile} pixels in the captured image sequence. Disregarding regions of the scene that spontaneously change -- e.g., blinking lights or digital screens -- pixel differences between images can be decomposed into the products of scene motion, illuminant motion, camera rotation, and depth parallax. Separating these sources of motion has been a long-standing challenge in vision~\cite{vogel2013piecewise,teed2021raft} as this is a fundamentally ill-conditioned problem; in typical settings, scene and camera motion are geometrically equivalent~\cite{hartley2003multiple}. One response to this problem is to disregard effects other than camera motion, which can yield high-quality motion estimates for static, mostly-lambertian scenes~\cite{yu20143d,im2015high,chugunov2023shakes}. This can be represented as
\begin{equation}\label{eq:sample_depth}
I(u,v,t) = [R,G,B] = f(\bm{\pi}\bm{\pi}_t^{-1}(u,v)),
\end{equation}
where $I(u,v,t)$ is a frame from the burst stack captured at time $t$ and sampled at image coordinates $u,v\in[0,1]$. Operators $\bm{\pi}$ and $\bm{\pi}_t$ perform 3D reprojection on these coordinates to transform them from time $t$ to the coordinates of a reference image model $f(u,v) \rightarrow [R,G,B]$. To account for other sources of motion, layer separation approaches such as \cite{kasten2021layered, nam2022neural} estimate a generic flow model $\Delta u, \Delta v = g(u,v,t)$ to re-sample the image model
\begin{equation}\label{eq:sample_flow}
I(u,v,t) = f(u + \scalebox{0.7}{$\Delta$} u, v + \scalebox{0.7}{$\Delta$} v).
\end{equation}
However, this parametrization introduces an overfitting risk, the consequences of which are illustrated in Fig.~\ref{fig:flow_representations}, as $g(u,v,t)$ and $f(u,v)$ can now act as a generic video encoder~\cite{li2023dynibar}. To combat this, methods often employ a form of gradient penalty such as total variation loss~\cite{nam2022neural}. That is
\begin{align}
    \mathcal{L}_{\text {TVFlow }}=\sum\left\|J_{g}(u, v, t)\right\|_1,\nonumber
\end{align}
where $J_g(u, v, t)$ is the Jacobian of the flow model. During training, this can prove computationally expensive, however, as now each sample requires its local neighborhood to be evaluated to numerically estimate the Jacobian, or a second gradient pass over the model. In both cases, a large number of operations are spent to limit the reconstruction of high frequency spatial and temporal content.

\noindent\textbf{Formulation.}\hspace{0.1em} We propose a neural spline field (NSF) model of flow which provides strong controls on reconstruction directly through its parametrization. This model splits flow evaluation into two components
\begin{equation}\label{eq:spline_field}
    \Delta u, \Delta v = g(u,v,t) = S(t; \mathbf{P} = h(u,v)).
\end{equation}
Here $h(u,v)$ is the NSF, a network which maps image coordinates to a set of spline control points $\mathbf{P}$. Then, to estimate flow for a frame at time $t$ in the burst stack, we evaluate the spline at $S(t;\mathbf{P})$. We select a cubic Hermite spline
\begin{align}\label{eq:cubic_spline}
    S(t,\mathbf{P}) &= (2t_r^3 - 3t_r^2 + 1) \mathbf{P}_{\lfloor t_s \rfloor} + (-2t_r^3 + 3t_r^2) \mathbf{P}_{\lfloor t_s \rfloor + 1} \nonumber\\
    &+ (t_r^3 - 2t_r^2 + t_r)(\mathbf{P}_{\lfloor t_s \rfloor} - \mathbf{P}_{\lfloor t_s \rfloor - 1})/2 \nonumber \\ 
      &+ (t_r^3 - t_r^2)(\mathbf{P}_{\lfloor t_s \rfloor + 1} - \mathbf{P}_{\lfloor t_s \rfloor})/2\nonumber \\
    t_r &= t_s - \lfloor t_s \rfloor, \quad t_s = t \cdot |\mathbf{P}|,
\end{align}
as it guarantees continuity in time with respect to its zeroth, first, and second derivatives and allows for fast local evaluation -- in contrast to B\'ezier curves~\cite{chugunov2023shakes} which require recursive calculations. We emphasize that the use of splines in graphics problems is {extensive}~\cite{farin2002curves}, and that there are many alternate candidate functions for $S(t,\mathbf{P})$. E.g., if the motion is expected to be a straight line, a piece-wise linear spline with $|\mathbf{P}|=2$ control points would insure this constraint is satisfied irregardless of the outputs of the NSF.

\begin{figure}[t]
    \centering
    \includegraphics[width=\linewidth]{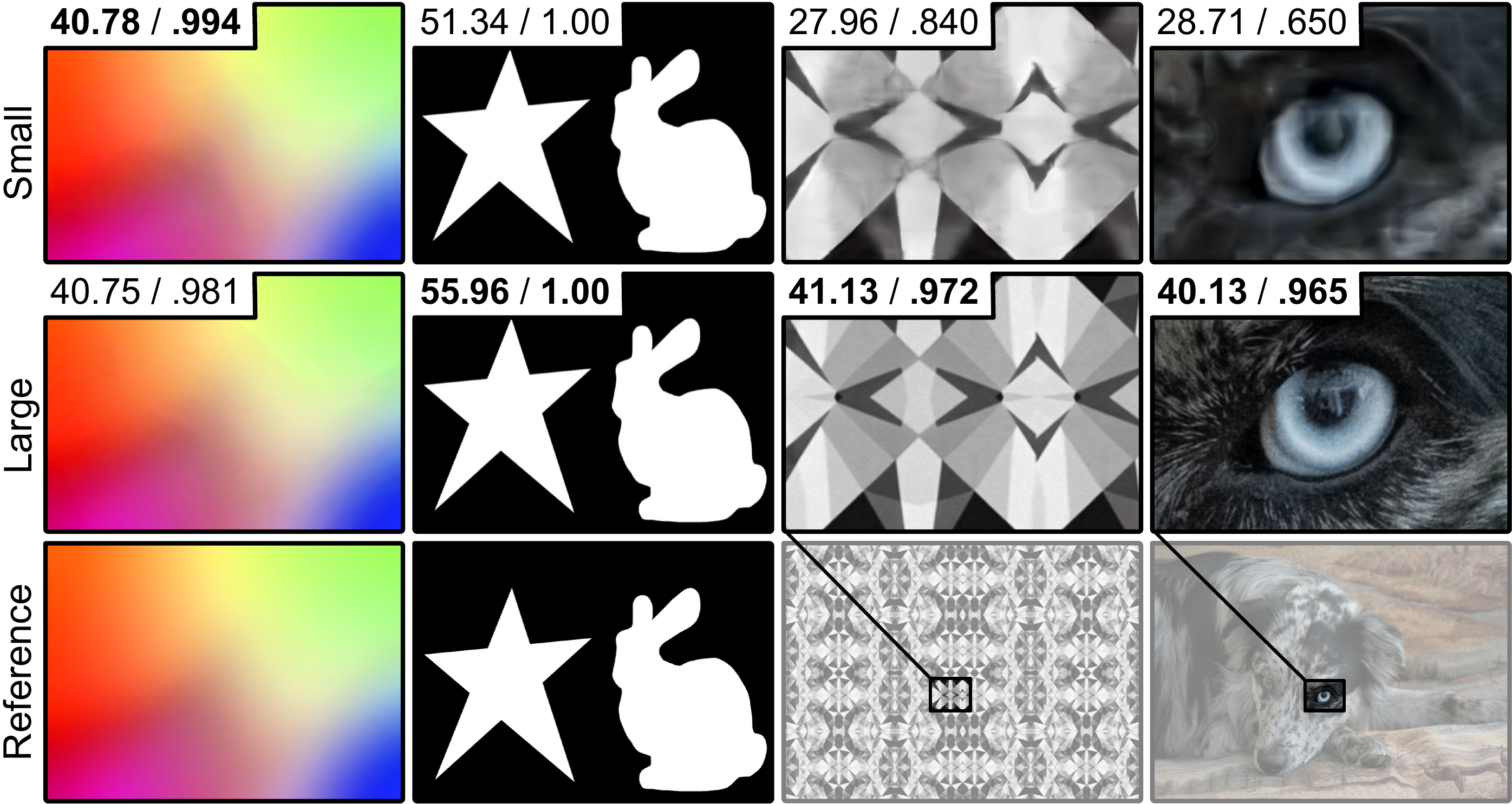}
    \vspace*{-1.5em}
    \caption{Image fitting results coordinate networks with \textit{Small} ($\mathrm{L}^\gamma{=}8$) and \textit{Large} ($\mathrm{L}^\gamma{=}16$) multi-resolution hash encodings and identical other parameters; PSNR/SSIM values inset top-left. Unlike a traditional band-limited representation~\cite{yang2022polynomial}, the \textit{Small} resolution network is able to fit both low-frequency smooth gradients and sharp edge mask images, but fails to fit a high density of either. This makes it a promising candidate representation for scene flow and alpha mattes which are comprised of smooth gradients and a limited number of object edges. }
    \label{fig:neural_fitting}
    \vspace{-1.5em}
\end{figure}

\noindent Where the choice of $S(t,\mathbf{P})$ and $|\mathbf{P}|$ determines the temporal behavior of flow, $h(u,v)$ controls its spatial properties. While our method, in principle, is not restricted to a specific spatial encoding function, we adopt the multi-resolution hash encoding $\gamma(u,v)$ presented in M\"uller et al.~\cite{muller2022instant}
\begin{align}\label{eq:multires-hash}
    h(u,v) &= \mathbf{h}(\gamma(u,v;\, \mathrm{params}_\gamma);\, \theta)\nonumber \\
    \mathrm{params}_\gamma &= \left\{\mathrm{B}^\gamma, \mathrm{S}^\gamma, \mathrm{L}^\gamma, \mathrm{F}^\gamma, \mathrm{T}^\gamma\right\},    
\end{align}
as it allows for fast training and strong  spatial controls given by its encoding parameters $\mathrm{params}_\gamma$: base grid resolution $\mathrm{B}^\gamma$, per level scale factor $\mathrm{S}^\gamma$, number of grid levels $\mathrm{L}^\gamma$, feature dimension $\mathrm{F}^\gamma$, and backing hash table size $\mathrm{T}^\gamma$. Here, $\mathbf{h}(\gamma(u,v);\theta)$ is a multi-layer perceptron (MLP)~\cite{hornik1989multilayer} with learned weights $\theta$. Illustrated in Fig.~\ref{fig:neural_fitting} with an image fitting example, the number of grid levels $\mathrm{L}^\gamma$ -- which, with a fixed $\mathrm{S}^\gamma$, sets the maximum grid resolution -- provides controls on the maximum ``spatial complexity" of the output while still permitting accurate reconstruction of image edges.

\begin{figure*}[t!]
    \centering
    \includegraphics[width=\linewidth]{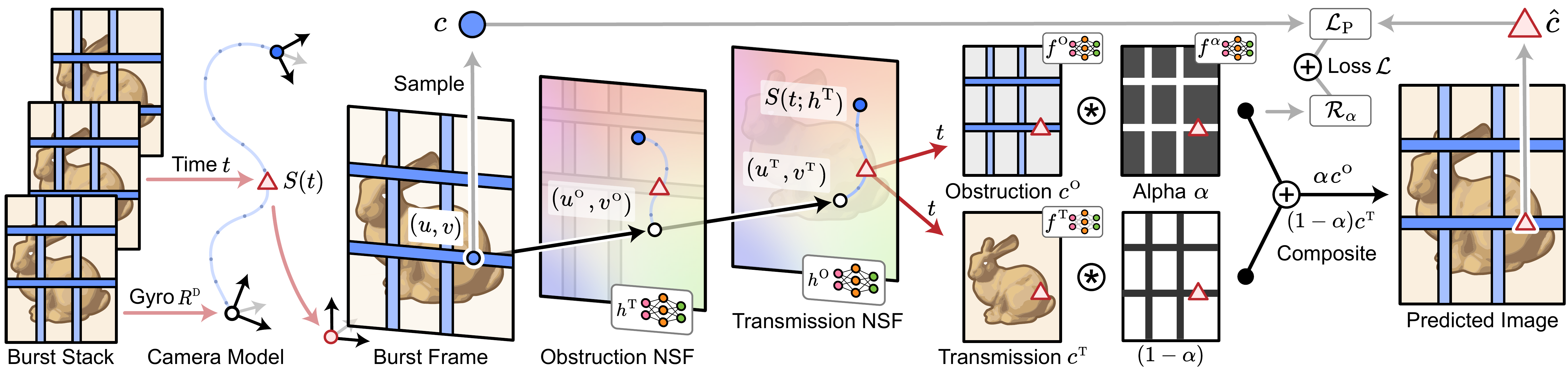}
    \caption{We model an input image sequence as the alpha composition of a \textit{transmission} and \textit{obstruction} plane. Motion in the scene is expressed as the product of a rigid camera model, which produces global rotation and translation, and two neural spline field models, which produce local flow estimates for the two layers. Trained to minimize photometric loss, this model separates content to its respective layers.}
    \label{fig:method}
    \vspace{-1.5em}
\end{figure*}

\subsection{Projective Model of Burst Photography} 
\noindent\textbf{Motivation.}\hspace{0.1em} With a flow model $g(u,v,t)$, and a canonical image representation $f(u,v)$ in hand, we theoretically have all the components needed to model an arbitrary image sequence~\cite{kasten2021layered,nam2022neural}. However, handheld burst photography does \textit{not} produce arbitrary image sequences; it has well-studied photometric and geometric properties~\cite{chugunov2023shakes,chugunov2022implicit,ha2016high,wronski2019handheld}. This, in combination with the abundance of physical metadata such as gyroscope values and calibrated intrinsics available on modern smartphone devices~\cite{chugunov2023shakes}, provides strong support for a physical model of image formation.  \\
\noindent\textbf{Formulation.}\hspace{0.1em} We adopt a forward model similar to traditional multi-planar imaging~\cite{hartley2003multiple}. We note that this departs from existing work~\cite{chugunov2022implicit,chugunov2023shakes}, which employs a backward projection camera model -- ``splatting" points from a canonical representation to locations in the burst stack. A multi-plane imaging model allows for both simple composition of multiple layers along a ray -- a task for which backward projection is not well suited -- and fast calculation of ray intersections without the ray-marching needed by volumetric representations like NeRF~\cite{mildenhall2020nerf}. For simplicity of notation, we outline this model for a single projected ray below. We also illustrate this process in Fig.~\ref{fig:method}. Let
\begin{align}
    c = [\mathrm{R},\mathrm{G},\mathrm{B}]^\top = I(u,v,t)
\end{align}
be a colored point sampled at time $t$ in the burst stack at image coordinates  $u,v\in [0,1]$. Note that these coordinates are relative to the camera pose at time $t$; for example $(u,v)=(0,0)$ is always the top left corner of the image. To project these points into world space we introduce camera translation $T(t)$ and rotation $R(t)$ models
\begin{align}\label{eq:translation_rotation}
    T(t) &= S(t,\mathbf{P}^\textsc{t}), \quad R(t) = R^\textsc{d}(t) + \eta_\textsc{r}S(t,\mathbf{P}^\textsc{r})\nonumber \\
    \mathbf{P}^\textsc{t}_i &= 
    \left[\arraycolsep=2.0pt
    \begin{array}{c}
    x \\
    y \\
    z \\
    \end{array}\right], \quad \mathbf{P}^\textsc{r}_i = \left[\begin{array}{ccc}
0 & -r^z & r^y \\
r^z & 0 & -r^x \\
-r^y & r^x & 0
\end{array}\right].
\end{align}
Here $S(t,\mathbf{P})$ is the same cubic spline model from Eq.~\eqref{eq:cubic_spline}, evaluated element-wise over the channels of $\mathbf{P}$. We note there are \textit{no coordinate networks} employed in these models. Translation $T(t)$ is learned from scratch, $\mathbf{P}^\textsc{t}$ initialized to all-zeroes. Rotation $R(t)$ is learned as a small-angle approximation offset~\cite{im2015high} to device rotations $R^\textsc{d}(t)$ recorded by the phone's gyroscope -- or alternatively, the identity matrix if such data is not available. With these two models, and calibrated intrinsic matrix $K$ from the camera metadata, we now generate a ray with origin $O$ and direction $D$ as
\begin{align}\label{eq:ray_generation}
    O \,{=}\, \left[\arraycolsep=2.0pt
    \begin{array}{c}
    O_x \\
    O_y \\
    O_z \\
    \end{array}\right] \,{=}\, T(t), \, \, D \,{=}\, \left[\arraycolsep=2.0pt
    \begin{array}{c}
    D_x \\
    D_y \\
    1 \\
    \end{array}\right] \,{=}\, \frac{R(t)K^{-1}}{D_z} \left[\arraycolsep=2.0pt
    \begin{array}{c}
    u \\
    v \\
    1 \\
    \end{array}\right],
\end{align}
where $D$ is normalized by its z component. We define our transmission and obstruction image planes as $\Pi^\textsc{t}$ and $\Pi^\textsc{o}$, respectively. As XY translation of these planes conflicts with changes in the camera pose, we lock them to the z-axis at depth $\Pi_z$ with canonical axes $\Pi_u$ and $\Pi_v$. Thus, given ray direction $D$ has a z-component of 1, we can calculate the ray-plane intersection as $Q = O + (\Pi_z - O_z)D$ and project to plane coordinates
\begin{align}\label{eq:plane_projection}
    u^{\scalebox{0.5}{$\Pi$}},v^{\scalebox{0.5}{$\Pi$}} = \langle Q, \, \Pi_u \rangle / (\Pi_z - O_z),\, \langle Q, \, \Pi_v \rangle / (\Pi_z - O_z),
\end{align}
scaled by ray length to preserve uniform spatial resolution. Let $u^\textsc{t},v^\textsc{t}$ and $u^\textsc{o},v^\textsc{o}$ be the intersection coordinates for the transmission and obstruction plane, respectively. We alpha composite these layers along the ray as
\begin{align}\label{eq:compositing}
\hat{c} &= (1-\alpha)c^\textsc{t} + \alpha c^\textsc{o}\nonumber \\
c^\textsc{t} &= f^\textsc{t}(u^\textsc{t} \,{+}\, \scalebox{0.7}{$\Delta$} u^\textsc{t}, v^\textsc{t} \,{+}\, \scalebox{0.7}{$\Delta$} v^\textsc{t}),\, \scalebox{0.7}{$\Delta$} u^\textsc{t}, \scalebox{0.7}{$\Delta$} v^\textsc{t} = S(t; h^\textsc{t}(u^\textsc{t}, v^\textsc{t})) \nonumber \\
c^\textsc{o} &\,{=}\, f^\textsc{o}(u^\textsc{o} \,{+}\, \scalebox{0.7}{$\Delta$} u^\textsc{o}, v^\textsc{o} \,{+}\, \scalebox{0.7}{$\Delta$} v^\textsc{o}),\, \scalebox{0.7}{$\Delta$} u^\textsc{o} \ucomma \scalebox{0.7}{$\Delta$} v^\textsc{o} =  S(t; h^\textsc{o}(u^\textsc{o}\ucomma v^\textsc{o}))\nonumber \\
\alpha&=\sigma(\tau_{\sigma} f^\alpha(u^\textsc{o} \,{+}\, \scalebox{0.7}{$\Delta$} u^\textsc{o},v^\textsc{o} \,{+}\, \scalebox{0.7}{$\Delta$} v^\textsc{o})),
\end{align}
where $\hat{c}$ is the composite color point, the weighted sum by $\alpha$ of the transmission color $c^\textsc{t}$ and obstruction color $c^\textsc{o}$. Each is the output of an image coordinate network $f(u,v)$ sampled at points offset by flow from an NSF $h(u,v)$. The sigmoid function $\sigma\,{=}\,1/(1+e^{-x})$ with temperature $\tau_\sigma$ controls the transition between opaque $\alpha\,{=}\,1$ and partially translucent $\alpha\,{=}\,0.5$ obstructions. This proves particularly helpful for learning hard occluders -- e.g., a fence -- where large $\tau_\sigma$ creates a steep transition between $\alpha\,{=}\,0$ and $\alpha\,{=}\,1$, which discourages $f^\alpha(u,v)$ from mixing content between layers.
\subsection{Training Procedure}
\begin{figure}[t]
    \centering
    \includegraphics[width=\linewidth]{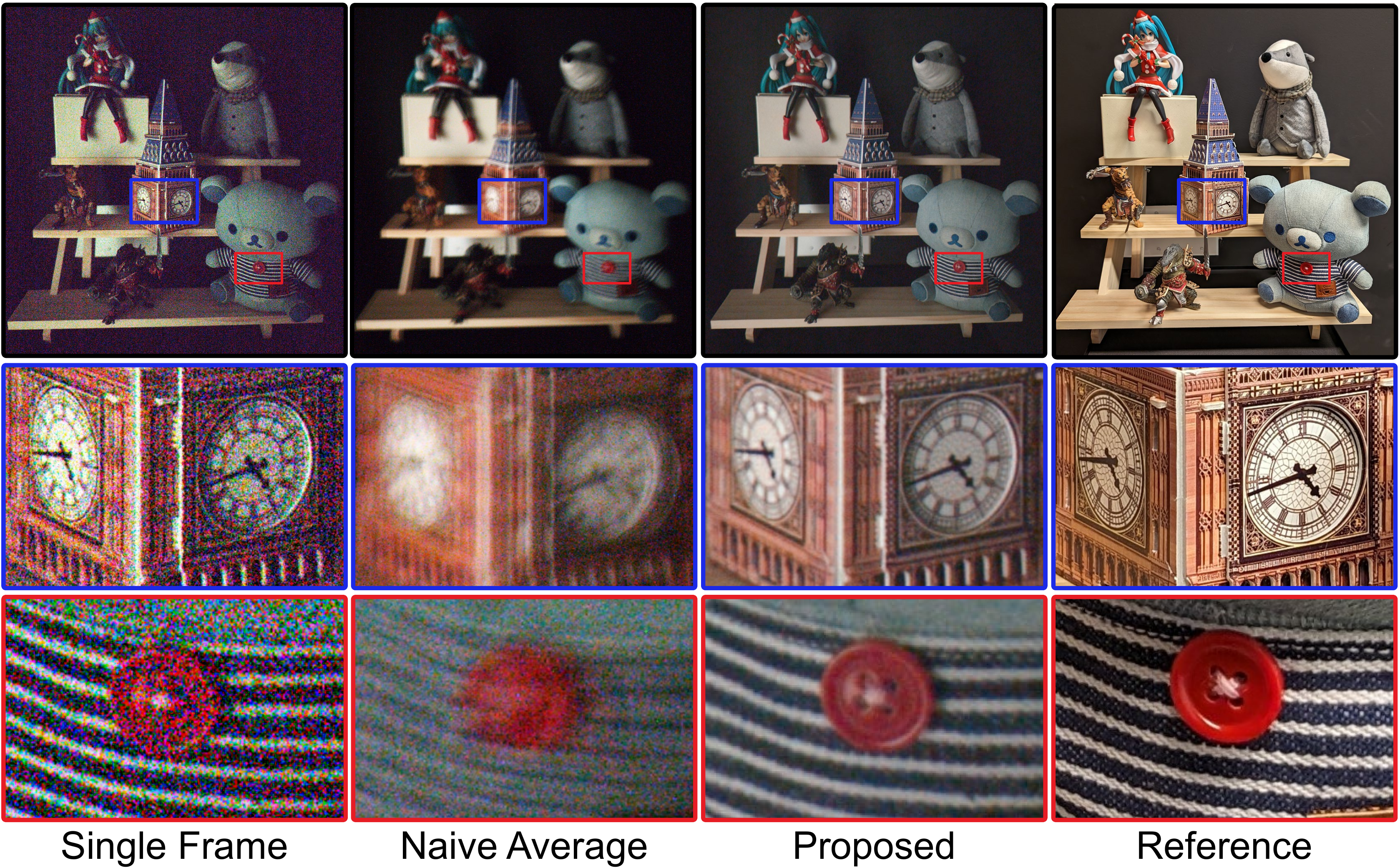}
    \vspace{-1.5em}
    \caption{Reconstruction results for noisy, low-light conditions; exposure time 1/30, ISO 5000. The proposed model is able to robustly merge frames into a denoised image representation.}
    \label{fig:lowlight}
    \vspace{-1.5em}
\end{figure}

\noindent\textbf{Losses.}\hspace{0.1em} Given all the components of our model are fully differentiable, we train them end-to-end via stochastic gradient descent. We define our loss  function $\mathcal{L}$ as
\begin{align}
    \mathcal{L} &= \mathcal{L}_\textsc{p} + \eta_\alpha\mathcal{R}_\alpha \label{eq:loss1}\\
    \mathcal{L}_\textsc{p} &= |(c - \hat c)/(\mathrm{sg}(c) + \epsilon)|, \quad  \mathcal{R}_\alpha = |\alpha|, \label{eq:loss2}\nonumber
\end{align}
where $\mathcal{L}_\textsc{p}$ is a relative photometric reconstruction loss~\cite{mildenhall2022nerf,chugunov2023shakes}. Demonstrated in Fig.~\ref{fig:lowlight}, when combined with linear RAW input data, this loss proves robust in noisy imaging settings~\cite{mildenhall2022nerf}. Thus, we select it for in-the-wild scene reconstruction with unknown conditions. The regularization $\mathcal{R}_\alpha$ with weight $\eta_\alpha$ controls the total contribution of the obstruction layer, discouraging it from duplicating content represented by the transmission layer. 
\begin{figure*}[t!]
    \centering
    \includegraphics[width=\linewidth]{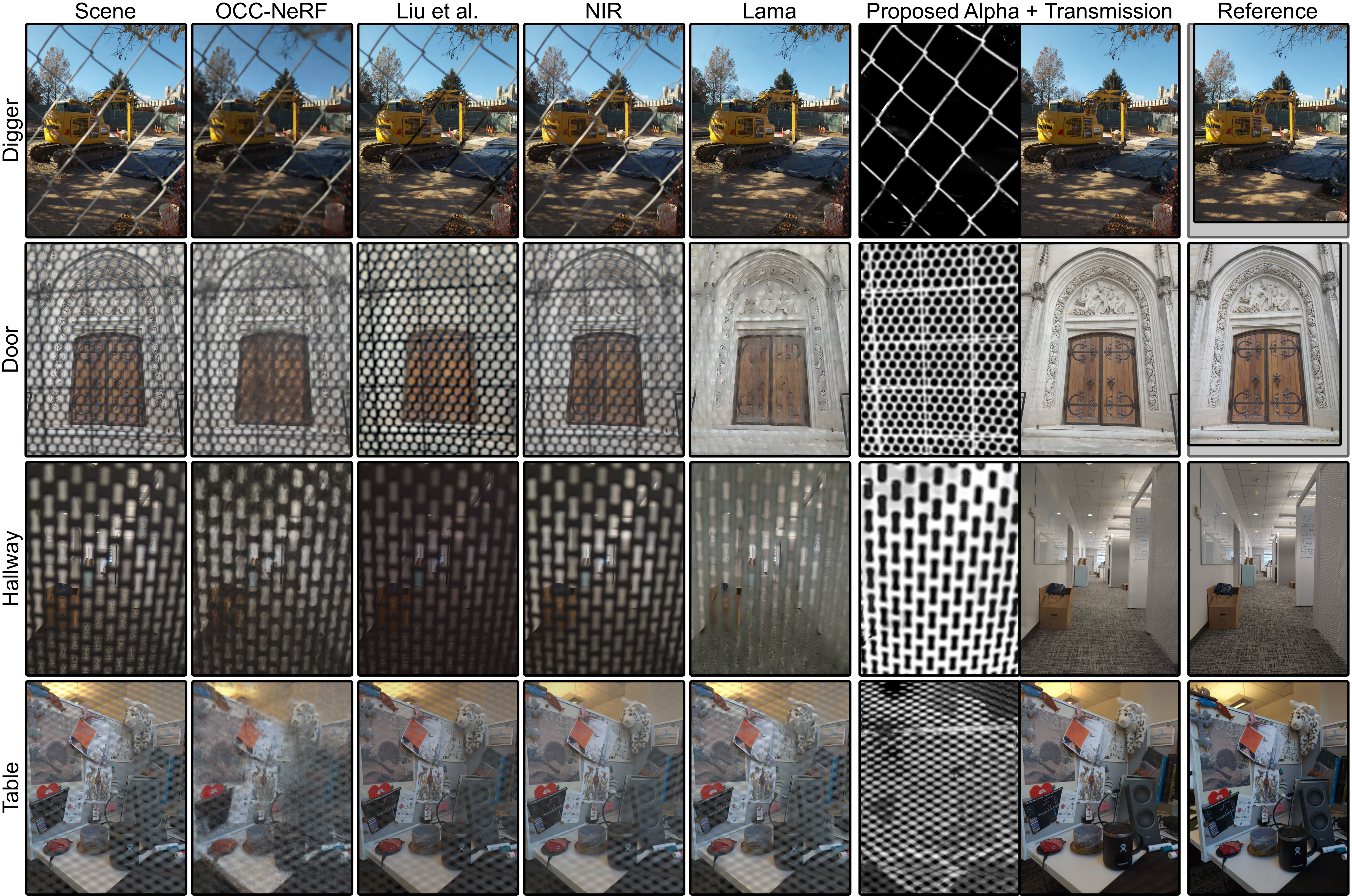}
    \vspace*{-1.5em}
    \caption{Occlusion removal results and estimated alpha maps for a set of captures with reference views, with comparisons to single image, multi-view, and NeRF fitting approaches. See video materials for visualization of input data and scene fitting.}
    \label{fig:results_occlusion}
    \vspace{-1.5em}
\end{figure*}

\noindent\textbf{Training.} Given the high-dimensional problem of jointly solving for camera poses, image layers, and neural spline field flows, we turn to coarse-to-fine optimization to avoid low-quality local minima solutions. During training, we mask the multi-resolution hash encodings $\gamma(u,v)$ input into our image, alpha, and flow networks as
\begin{align}
    \gamma_i(u,v) &= 
\begin{cases} 
\gamma_i(u,v) & \text{if } \,\, i/|\gamma| < 0.4 + 0.6(\text{sin\_epoch}) \\
0 & \text{if } \,\, i/|\gamma| > 0.4 + 0.6(\text{sin\_epoch})
\end{cases}\nonumber\\
\text{sin\_epoch}&=\mathrm{sin}(\text{epoch}/\text{max\_epoch}),
\end{align}
activating higher resolution grids as training progresses. This strategy results in less noise accumulated during early training as spurious high-resolution features do not need to be ``unlearned"~\cite{chugunov2023shakes,li2023neuralangelo} during later stages of refinement. 

\vspace{-0.5em}
\section{Applications}\label{sec:results}
\noindent\textbf{Data Collection.} To collect burst data we modify the open-source Android camera capture tool \href{https://github.com/Ilya-Muromets/Pani}{Pani} to record continuous streams of RAW frames and sensor metadata. During capture, we lock exposure and focus settings to record a 42 frame, two-second ``long-burst" of 12-megapixel images, gyroscope measurements, and camera metadata. We refer the reader to Chugunov et al.~\cite{chugunov2023shakes} for an overview of the long-burst imaging setting and its geometric properties. We capture data from a set of Pixel 7, 7-Pro, and 8-Pro devices, with no notable differences in overall reconstruction quality or changes in the training procedure required. We train our networks directly on Bayer RAW data, and apply device color-correction and tone-mapping for visualization.

\noindent\textbf{Implementation Details.} During training, we perform stochastic gradient descent on $\mathcal L$ for batches of $2^{18}$ rays per step for $6000$ steps with the Adam optimizer~\cite{kingma2014adam}. All networks use the multi-resolution hash encoding described in Eq.~\eqref{eq:multires-hash}, implemented in tiny-cuda-nn~\cite{muller2021real}. Trained on a single Nvidia RTX 4090 GPU, our method takes approximately \textit{3 minutes} to fit a full 42-frame image sequence. All networks have a base resolution $B^\gamma{=}4$, and scale factor $S^\gamma{=}1.61$, but while flow networks $h^\textsc{t}$ and $\textsc{o}$ are parameterized with a low number of grid levels $L^\gamma{=}8$, networks which represent high frequency content have $L^\gamma{=}12$ or $L^\gamma{=}16$ levels. These settings are task-specific, and full implementation details and results for short (4-8 frame) image bursts are included in the Supplementary Material.

%% file: 3_results.tex
\begin{figure}[h!]
    \centering
    \includegraphics[width=\linewidth]{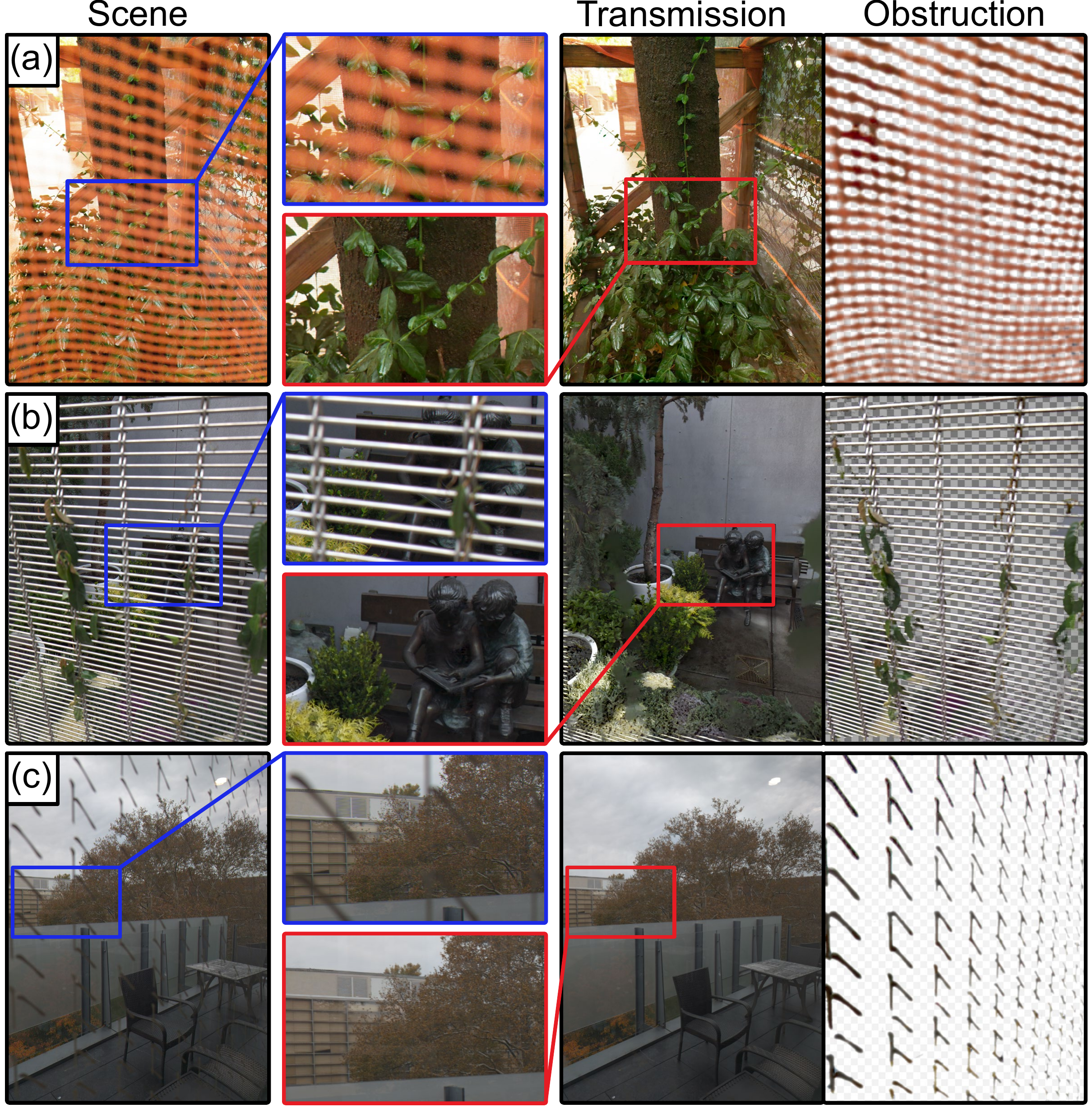}
    \vspace*{-2em}
    \caption{Layer separation results in unique real-world cases enabled by  our generalizable two-layer image model: (a) orange planter, (b) fenced garden, (c) stickers on balcony glass.}
    \label{fig:misc_scenes}
    \vspace{-2em}
\end{figure}
\begin{figure*}[t!]
    \centering
    \includegraphics[width=\linewidth]{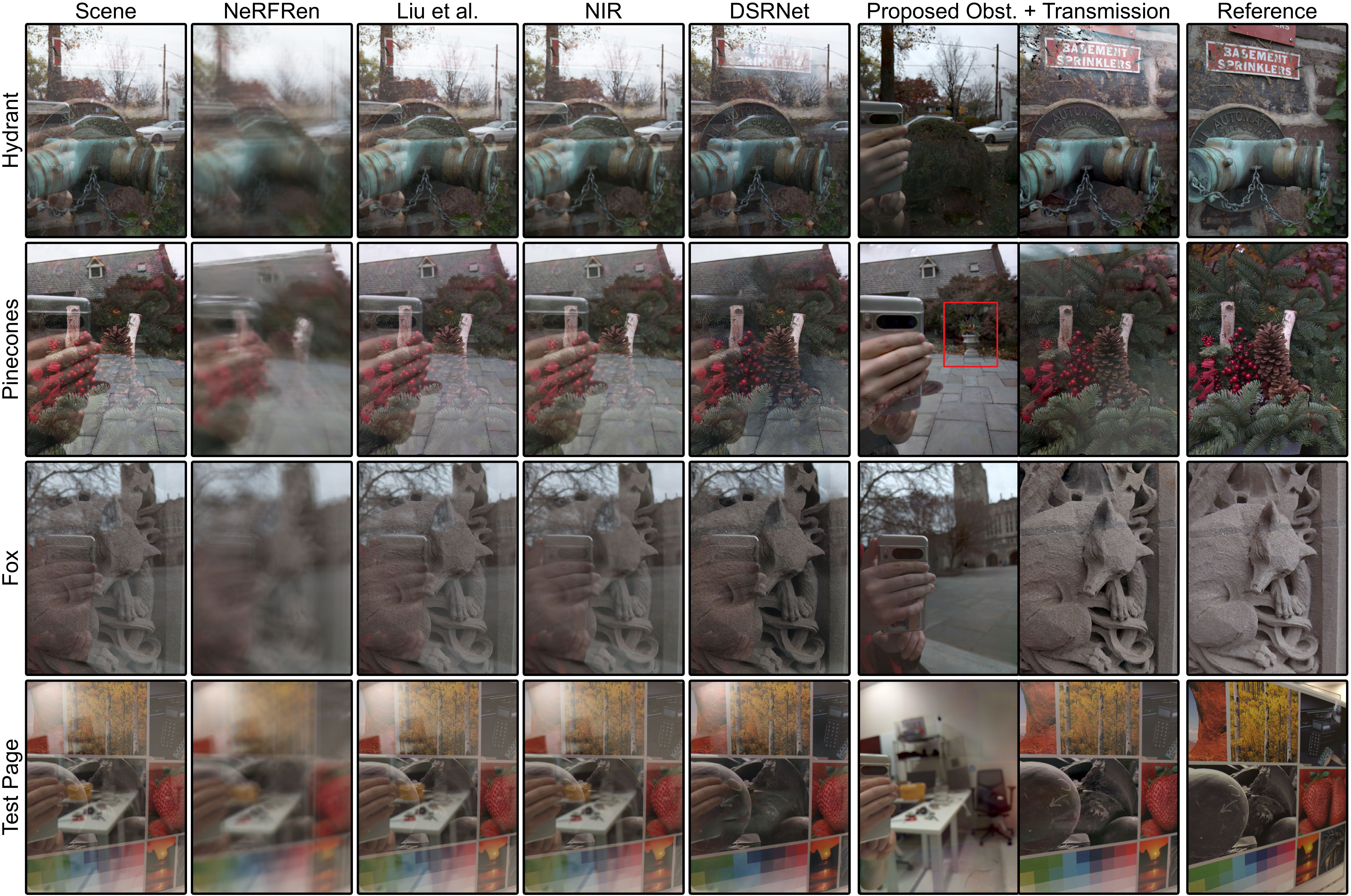}
    \vspace*{-1.5em}
    \caption{Reflection removal results and estimated alpha maps for a set of captures with reference views, with comparisons to single image, multi-view, and NeRF fitting approaches. See video materials for visualization of input data and scene fitting.}
    \label{fig:results_reflection}
    \vspace{-1.5em}
\end{figure*}
\noindent\textbf{Occlusion Removal.}\hspace{0.1em} Initializing the obstruction plane closer to the camera than the transmission plane, that is $\Pi^\textsc{o}_z < \Pi^\textsc{t}_z$, we find that the $f^\textsc{o}(u,v)$ naturally reconstructs foreground content in the scene. Given a scene with content hidden behind a foreground occluder -- e.g., imaging through a fence --  we can then perform occlusion removal with the proposed method by setting $\alpha=0$. We report results in Fig.~\ref{fig:results_occlusion} for a set of captures collected with reference views using a tripod-mounted occluder. We compare here to the multiview plus learning method presented in \textit{Liu et al.}~\cite{liu2020learning}, the neural radiance field approach \textit{OCC-NeRF}~\cite{zhu2023occlusion}, the flow + homography neural image model \textit{NIR}~\cite{nam2022neural}, and the single image inpainting method \textit{Lama}~\cite{suvorov2021resolution} as these methods demonstrate a broad range of techniques for occlusion detection and removal with varying assumptions on camera motion. We find that in this small baseline burst photography setting, existing multi-view methods fail to achieve meaningful occlusion removal; as the occluder maintains a high level of self-overlap for the whole image sequence.
\begin{figure}[h!]
    \centering
    \includegraphics[width=\linewidth]{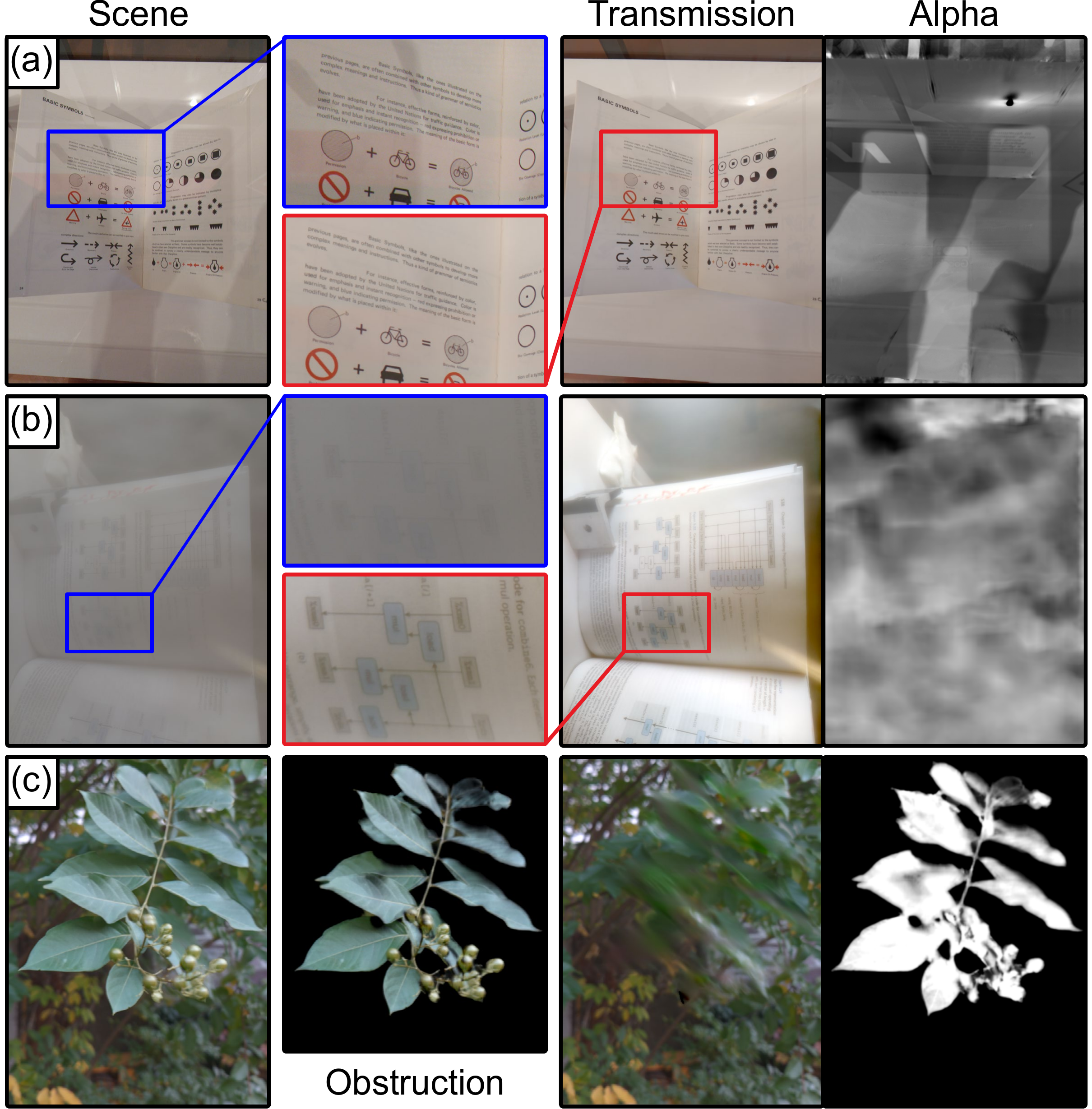}
    \vspace*{-2em}
    \caption{Layer separation results for additional example applications: (a) shadow removal, (b) image dehazing, and (c)  video motion segmentation (see \href{https://light.princeton.edu/publication/nsf}{video materials} for visualization).}
    \label{fig:in_the_wild}
    \vspace{-2em}
\end{figure}
\begin{figure*}[t!]
    \centering
    \includegraphics[width=\linewidth]{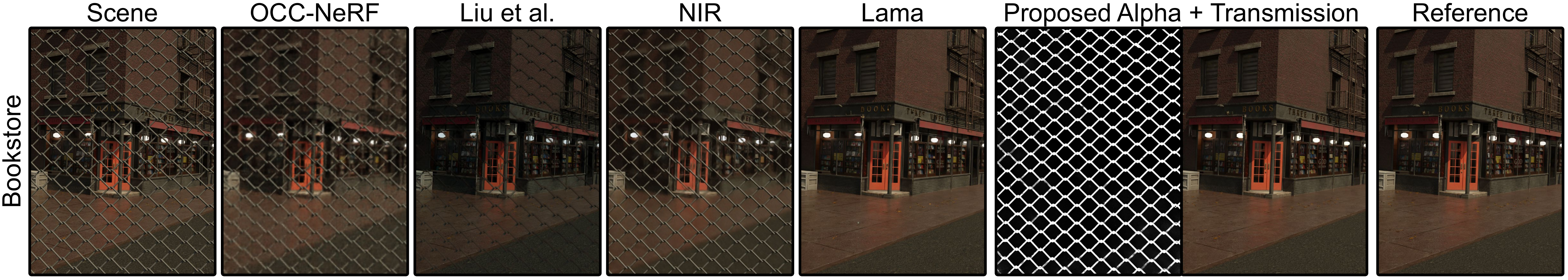}
        \resizebox{\linewidth}{!}{
     \begin{tabular}[b]{lccccclccccc}
     \midrule
          \textbf{Occlusion} & OCC-NeRF & Liu et al. & NIR & Lama & Proposed & \textbf{Occlusion} & OCC-NeRF & Liu et al. & NIR & Lama & Proposed\\
	\midrule
	\midrule
    \textit{Bookstore} & 21.34$/$0.467 & 24.57$/$0.716 & 20.44$/$0.459 & 28.72$/$0.824 & \textbf{35.59/0.929} & \textit{Window}  & 14.22$/$0.608 & 14.03$/$0.718 & 13.98$/$0.680 & 30.44$/$0.939 & \textbf{32.91/0.953}    \\
    \textit{Couch} & 13.37$/$0.658 & 13.09$/$0.708 & 13.09$/$0.690 & 26.86$/$0.927 & \textbf{31.65/0.963} & \textit{Station}   & 14.62$/$0.569 & 14.40$/$0.649 & 14.05$/$0.595 & 26.48$/$0.874 & \textbf{30.25/0.936}     \\
    \bottomrule
  \end{tabular}
  }
\includegraphics[width=\linewidth]{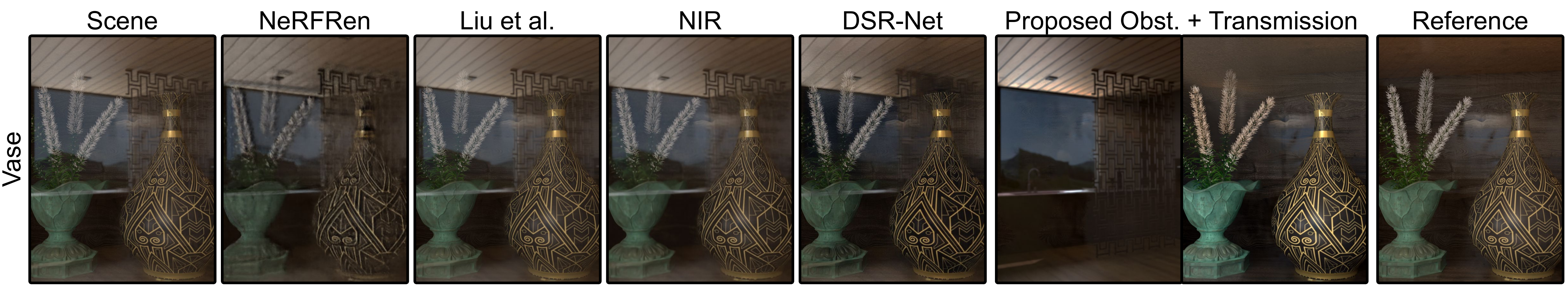}
 \resizebox{\linewidth}{!}{
     \begin{tabular}[b]{lccccclccccc}
     \midrule
          \textbf{Reflection} & NeRFReN & Liu et al. & NIR & DSR-Net & Proposed & \textbf{Reflection} & NeRFReN & Liu et al. & NIR & DSR-Net & Proposed\\
	\midrule
	\midrule
    \textit{Vase} & 18.40$/$0.688 & 18.53$/$0.849 & 18.52$/$0.771 & 19.43$/$0.835 & \textbf{27.20/0.942} & \textit{Barrels}  & 20.59$/$0.736 & 20.76$/$0.850 & 20.57$/$0.752 & 20.39$/$0.811 & \textbf{26.07/0.921}    \\
    \textit{Fish} & 20.05$/$0.744 & 24.32$/$0.904 & 19.78$/$0.783 & 23.51$/$0.875 & \textbf{25.47/0.824} & \textit{Mixer}   & 19.57$/$0.628 & 19.20$/$0.726 & 19.34$/$0.660 & 24.56$/$0.794 & \textbf{27.06/0.875}     \\
    \bottomrule
  \end{tabular}
  }
\vspace*{-1.5em}
    \caption{Qualitative and quantitative obstruction removal results for a set of 3D rendered scenes with paired ground truth, camera motion simulated from real measured hand shake data~\cite{chugunov2022implicit}. Evaluation metrics formatted as PSNR/SSIM.}
    \label{fig:results_synthetic}
    \vspace{-1.5em}
\end{figure*}
 While the single-image method, \textit{Lama} is able to in-paint occluded regions based on un-occluded content, it cannot faithfully recover lost details such as the carvings in the \textit{Door} scene. Furthermore, it does not produce an alpha matte, and rather \textit{requires a hand-annotated mask as input}. In contrast, our approach distills information from all input frames to accurately recover temporarily occluded content, and jointly produces a high-quality alpha matte. In Fig.~\ref{fig:in_the_wild} we present additional layer separation results for real in-the-wild scenes with complex occluders, which demonstrate the versatility of the obstruction image model $f^\textsc{o}(u,v)$. \\
\noindent\textbf{Reflection Removal.}\hspace{0.1em} We show in Fig.~\ref{fig:results_reflection} how by flipping the plane depths $\Pi^\textsc{o}_z > \Pi^\textsc{t}_z$,  our model is also able to separate reflected from transmitted content. Here, we compare again to \textit{Liu et al.}~\cite{liu2020learning} and \textit{NIR}~\cite{nam2022neural}, as well as the reflection-specific neural radiance approach \textit{NeRFReN}~\cite{guo2022nerfren} and single-image reflection removal network \textit{DSR-Net}~\cite{hu2023single}. Similarly to occlusion removal, we observe that given small-baseline inputs the multi-view methods fail to achieve meaningful layer separation, and \textit{NeRFRen} struggles to converge on a sharp reconstruction. Only \textit{DSR-Net} is able to suppress even small parts of the reflection such as the car in the \textit{Hydrant} scene. In contrast, the proposed method not only estimates nearly reflection-free transmission layers, but is also able to recover hidden content -- such as the flowerpot highlighted in \textit{Pinecones} -- in the reflection layer. \\
\noindent\textbf{Synthetic Validation.}\hspace{0.1em} To further validate our method, we construct a set of 3D scenes to render paired ground truth data, and provide quantitative and qualitative results in Fig.~\ref{fig:results_synthetic} and the supplementary material. These findings align with our findings from real-world captures, with significant PSNR and SSIM improvements across all scenes.\\
\noindent\textbf{Image Enhancement through Layer Separation.} In addition to occlusion and reflection removal, a wide range of other computational photography applications can be viewed through the lens of layer separation. We showcase several example tasks in Fig.~\ref{fig:misc_scenes}, including shadow removal, image dehazing, and video motion segmentation. The key relationship between all these tasks is that the two effects undergo different motion models -- e.g., photographer-cast shadows move with the cellphone, while the paper target stays static. By grouping color content with its respective motion model, $f^\textsc{t}(u,v)$ with $h^\textsc{t}(u,v)$ and $f^\textsc{o}(u,v)$ with $h^\textsc{o}(u,v)$, just as in the occlusion case, we can remove the effect by removing its image plane. Fig.~\ref{fig:misc_scenes} (c), which fits our two-layer model for an image sequence of a moving tree branch, also highlights that our method does not rely solely on camera motion. Scene motion itself can also be used as a mechanism for layer separation in image bursts, similar to approaches in video masking~\cite{lu2021omnimatte,kasten2021layered}.
\vspace{-0.5em}

%% file: 4_conclusion.tex
\vspace{-0.5em}\section{Discussion and Future Work}\vspace{-0.5em}
\noindent In this work, we present a versatile representation of burst photography built on a novel neural spline field model of flow, and demonstrate image fusion and obstruction removal results under a wide array of conditions. In future work, we hope this generalizable model can be tailored to specific layer separation and image fusion applications: \\
\noindent\textbf{Learned Features.} The proposed model relies entirely on photometric loss, with no visual priors on the scene. Learned features could help disambiguate content layers in areas without reliable parallax or motion information. \\ 
\noindent\textbf{Physical Priors.} Our generic image plus flow representation can accommodate task-specific modules for applications where there are known physical models, such as chromatic aberration removal or refractive index estimation. \\
\noindent\textbf{Beyond Burst Data.} There exist many other sources of multi-image data to which the method can potentially be adapted -- e.g., microscopes, telescopes, and light field, time-of-flight, or hyperspectral cameras.



%% file: supp_0_implementation.tex
\newpage
\section*{Supplementary Material}
\noindent In this supplementary material, we provide implementation details, additional results, ablation studies, and experimental analysis in support of the findings of the main text. The structure of this document is as follows:
\begin{itemize}
\itemsep0em 
    \item Section~\ref{sec:supp_implementation}: Details on data generation, model implementation, and training procedure.
    \item Section~\ref{sec:supp_results}:
    Additional obstruction removal results with comparison methods and synthetic validation. Analysis of challenging reconstruction settings.
    \item Section~\ref{sec:supp_experiments}: Additional analysis on manipulating model and training parameters. Includes reconstruction results for subsampled and short burst sequences. 
\end{itemize}
\setcounter{figure}{0}
\renewcommand*{\theHfigure}{chX.\the\value{figure}}
\setcounter{section}{0}
\renewcommand*{\theHsection}{chX.\the\value{section}}
\renewcommand\thesection{\Alph{section}}
\section{Implementation Details}
\label{sec:supp_implementation}
\begin{figure}[t]
    \centering
    \includegraphics[width=0.9\linewidth]{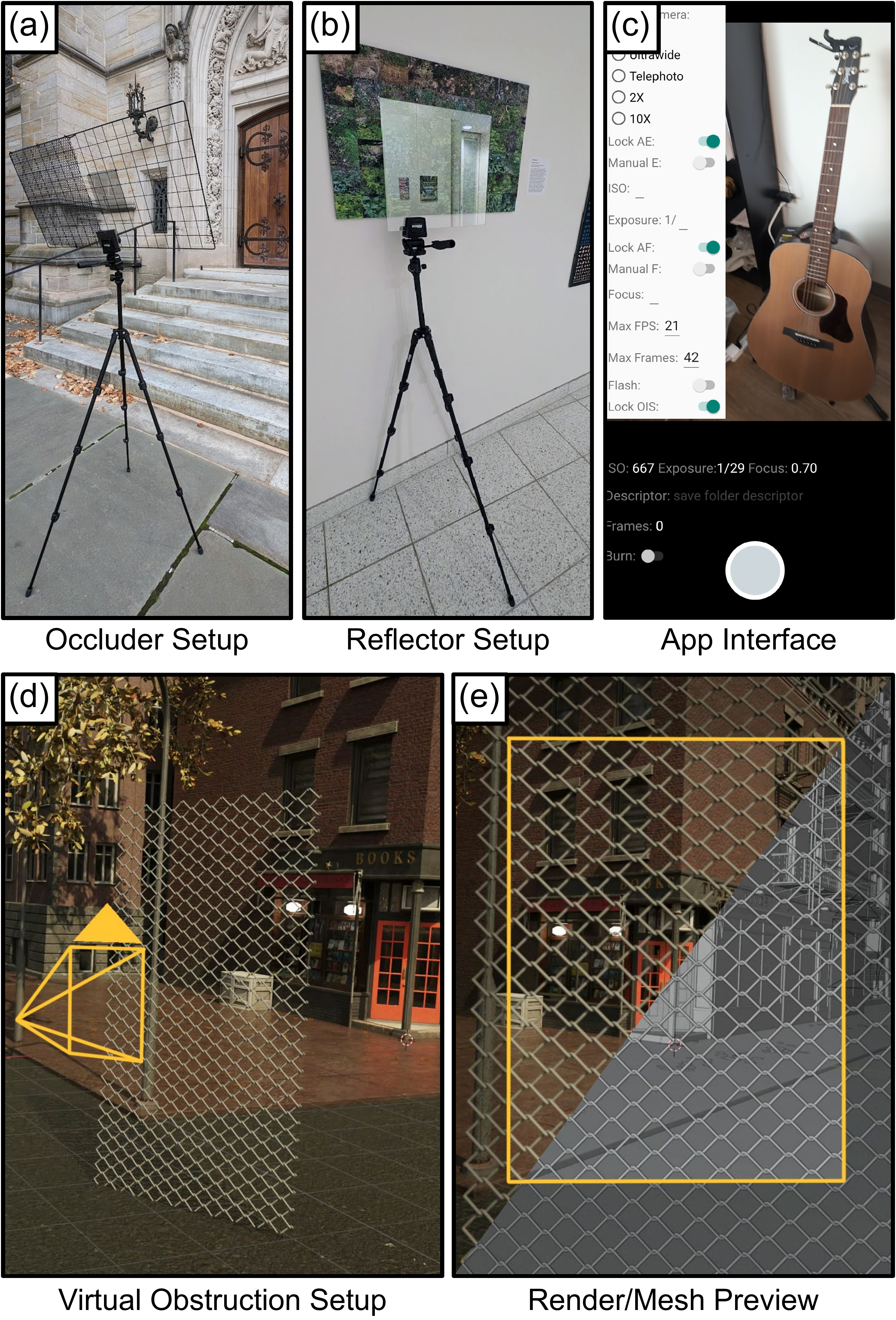}
    \caption{(a) Tripod-mounted occluder setup for capturing paired occlusion removal data. (b) Tripod-mounted reflector setup for capturing paired reflection removal data. (c) Capture app interface with the extended settings menu. (d-e) Example 3D scene with simulated occluder, camera frustum highlighted in orange.}
    \label{fig:sup_data_collection}
\end{figure}

\noindent\textbf{Data Acquisition}\hspace{0.1em} To acquire paired obstructed and unobstructed captures, we construct two tripod-mounted rigs as illustrated in Fig.~\ref{fig:sup_data_collection} (a-b). We begin by capturing a still of the scene without the obstruction, before rotating the tripod into position to capture a 42-frame obstructed long-burst~\cite{chugunov2022implicit} of 12-megapixel RAW frames. For accessible natural occluders, such as the fences in Fig.~\ref{fig:supp_results_occlusion}, we acquire reference views by positioning the phone at a gap in the occluder -- though this sometimes cannot perfectly remove the occluder as in the case of Fig.~\ref{fig:supp_results_occlusion} \textit{Pipes}. We collect data with our modified \href{https://github.com/Ilya-Muromets/Pani}{Pani} capture app, illustrated in Fig.~\ref{fig:sup_data_collection} (c), built on the Android camera2 API. During capture, we also record metadata such as camera intrinsics, exposure settings, channel color correction gains, tonemap curves, and other image processing and camera information during capture. We stream gyroscope and accelerometer measurements from on-board sensors as $\approx$100Hz, though we find accelerometer values to be highly unreliable for motion on the scale of natural hand tremor, and so disregard these measurements for this work. We apply minimal processing to the recorded 10-bit Bayer RAW frames -- only correcting for lens shading and BGGR color channel gains -- before splitting them into a 3-plane RGB color volume. We do not perform any further demosaicing on this volume, as these processes correlate local signal values, and instead input it directly into our model for scene fitting. For visualization, we apply the default color correction matrix and tone-curve supplied in the capture metadata.

\begin{figure}[t!]
    \centering
    \includegraphics[width=\linewidth]{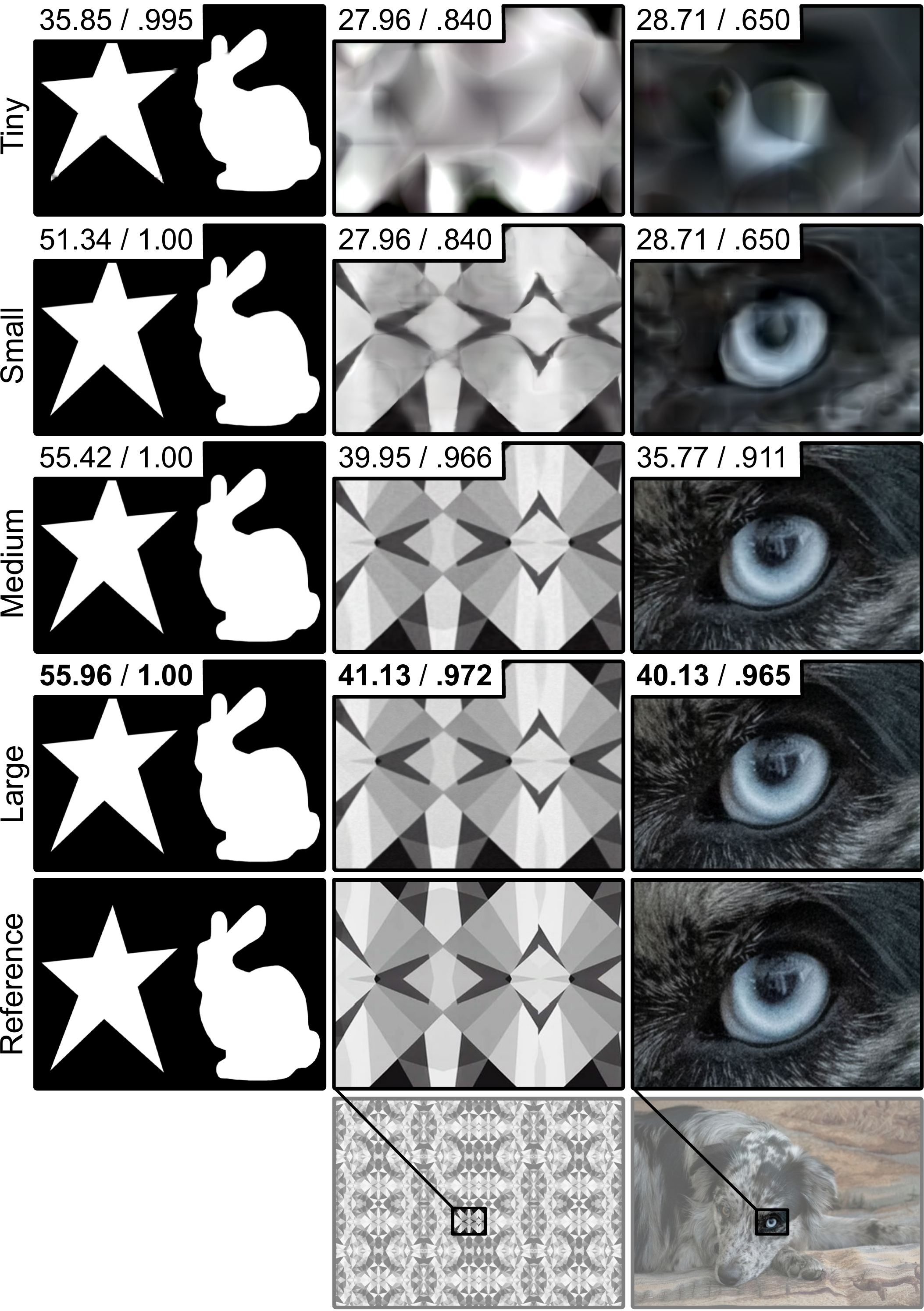}
    \caption{Image fitting results for network encoding configurations as described in Tab.~\ref{tab:network-sizes}, other training and network parameters held constant: 5-layer MLP coordinate networks, hidden dimension 64, ReLU activations. PSNR/SSIM values inset top-left.}
    \label{fig:supp_image_fitting}
    \vspace{-1.5em}
\end{figure}

\noindent\textbf{Synthetic Data Generation}\hspace{0.1em} Capturing aligned ground-truth data for obstruction removal is a long-standing problem in the field~\cite{wei2019single}, greatly exacerbated by the requirement in our setting of \textit{a sequence} of unstabilized frames with its base frame aligned to an unobstructed image. Thus, to generate data for further validation of our method, we turn to synthetic captures created through the 3D rendering software Blender. We select a set of 3D scenes from the \href{https://www.blenderkit.com/}{BlenderKit} asset library and manually tune their lighting and materials to reduce non-physical rendering artifacts, such as the replacement of point light sources with small radius spherical sources. We add \textit{occluder} and \textit{reflector} objects to these scenes in the form of fences, window gratings, and glass enclosures containing textured meshes. To generate realistic camera motion, we record samples of natural hand tremor with a pose-capture application built on the Apple ARKit library~\cite{chugunov2022implicit}. We then place cameras into the scene pointed at the synthetic obstructors, illustrated in Fig.~\ref{fig:sup_data_collection}, and offset their rotations and translations over time with the collected natural hand tremor data. We then render this data in the Cycles ray-tracing engine with 4000 samples per ray. Reconstruction results with this data are demonstrated in Fig.~\ref{fig:supp_occlusion_synthetic} and Fig.~\ref{fig:supp_reflection_synthetic}.\\
\noindent\textbf{Implementation Details}\hspace{0.1em} While the overarching model structure is held constant between all applications -- identical projection, image generation, and flow models for all tasks -- elements such as the neural spline field $h(u,v)$ encoding parameters $\mathrm{params}_\gamma$ can be tuned for specific tasks:
\begin{align}\label{eq:supp_multires-hash}
    h(u,v) &= \mathbf{h}(\gamma(u,v;\, \mathrm{params}_\gamma);\, \theta)\nonumber \\
    \mathrm{params}_\gamma &= \left\{\mathrm{B}^\gamma, \mathrm{S}^\gamma, \mathrm{L}^\gamma, \mathrm{F}^\gamma, \mathrm{T}^\gamma\right\}.   
\end{align}
\begin{table}[t!]
\centering
\begin{tabular}{cccccc}
\hline
 & base & scale & levels & feat.  & table \\
Size & $\mathrm{B}^\gamma$ & $\mathrm{S}^\gamma$ & $\mathrm{L}^\gamma$ & $\mathrm{F}^\gamma$ & $\mathrm{T}^\gamma$ \\\hline \hline
\textit{Tiny} (T) & 4 & 1.61 & \textbf{6} & 4 & \textbf{12} \\
\textit{Small} (S) & 4 & 1.61 & \textbf{8} & 4 & \textbf{14} \\
\textit{Medium} (M) & 4 & 1.61 & \textbf{12} & 4 & \textbf{16} \\
\textit{Large} (L) & 4 & 1.61 & \textbf{16} & 4 & \textbf{18} \\ \hline
\end{tabular}
\caption{Multi-resolution hash-table encoding parameters for different ``sizes'' of network, with larger encodings intended to fit higher-resolution data. Note that we only vary the number of grid levels $\mathrm{L}^\gamma$, and match the backing table size $\mathrm{T}^\gamma$ accordingly to avoid hash collisions. The base grid resolution $\mathrm{B}^\gamma$, grid per-level scale $\mathrm{S}^\gamma$, and feature encoding size $\mathrm{F}^\gamma$ are kept constant. }
\label{tab:network-sizes}
\vspace{1em}
\centering
\begin{tabular}{ccccccc}
\multicolumn{7}{l}{\textit{\textbf{occlusion removal}}:} \\
\hline
& flow $h$ & $|h|$ & rgb $f$ & $f^\alpha$  & depth $\Pi_z$ & $\eta_\alpha\mathcal{R}$ \\
\hline \hline
$Tr$: & T & 11 & L &   & 1.0 & \multirow{2}{*}{0.02} \\
$Ob$: & T & 11 & M & M  & 0.5 & \vspace{0.5em}\\ 
\multicolumn{7}{l}{\textit{\textbf{reflection removal}}:} \\
\hline
& flow $h$ & $|h|$ & rgb $f$ & $f^\alpha$  & depth $\Pi_z$ & $\eta_\alpha\mathcal{R}$ \\
\hline \hline
$Tr$: & T & 11 & L &   & 1.0 & \multirow{2}{*}{0.0} \\
$Ob$: & T & 11 & T & L  & 2.5 & \vspace{0.5em}\\ 
\multicolumn{7}{l}{\textit{\textbf{video segmentation}}:} \\
\hline
& flow $h$ & $|h|$ & rgb $f$ & $f^\alpha$  & depth $\Pi_z$ & $\eta_\alpha\mathcal{R}$ \\
\hline \hline
$Tr$: & S & 15 & L &   & 1.0 & \multirow{2}{*}{0.005$^\dagger$} \\
$Ob$: & S & 15 & L & M & 2.0 &  \vspace{0.5em}\\
\multicolumn{7}{l}{\textit{\textbf{shadow removal}}:} \\
\hline
& flow $h$ & $|h|$ & rgb $f$ & $f^\alpha$  & depth $\Pi_z$ & $\eta_\alpha\mathcal{R}$ \\
\hline \hline
$Tr$: & T & 11 & L &   & 1.0 & \multirow{2}{*}{0.0} \\
$Ob$: & T & 11 & T & M  & 2.0 &  \vspace{0.5em}\\
\multicolumn{7}{l}{\textbf{\textit{dehazing}:}} \\
\hline
& flow $h$ & $|h|$ & rgb $f$ & $f^\alpha$  & depth $\Pi_z$ & $\eta_\alpha\mathcal{R}$ \\
\hline \hline
$Tr$: & T & 11 & L &   & 1.0 & \multirow{2}{*}{-0.01} \\
$Ob$: & T & 11 & T & S  & 0.5 & \vspace{0.5em}\\ 
\multicolumn{7}{l}{\textit{\textbf{image fusion}}:} \\
\hline
& flow $h$ & $|h|$ & rgb $f$ & $f^\alpha$  & depth $\Pi_z$ & $\eta_\alpha\mathcal{R}$ \\
\hline \hline
$Tr$: & S & 31 & L &  & 1.0 & 0.0 \\
\end{tabular}
\caption{Network encoding, flow, and loss configurations used for several layer-separation applications, separated into rows individually defining transmission $Tr$ and obstruction $Ob$ layers. Encoding parameters are defined by the corresponding (T,S,M,L) row of Tab.~\ref{tab:network-sizes}. Flow size $|h|$ indicates the number of spline control points used for interpolation of the corresponding neural spline field $S(t,h(u,v))$. In the \textit{video segmentation} task, alpha loss with weight 0.005$^\dagger$ is \textit{segmentation} alpha loss $\alpha(1-\alpha)$ rather than $|\alpha|$.  }
\label{tab:application-configs}
\vspace{-1em}
\end{table}
By manipulating the parameters of Eq.~\ref{eq:supp_multires-hash} as defined in Tab.~\ref{tab:network-sizes} we construct four different ``sizes'' of network encodings to investigate the method behavior: \textit{Tiny}, \textit{Small}, \textit{Medium}, and \textit{Large}. Image fitting results in Fig.~\ref{fig:supp_image_fitting} illustrate what scale of features each of these configurations is able to reconstruct, with larger encoding reconstructing denser and higher-frequency content. Then, assembling together multiple image and flow networks with varying encoding sizes as defined in Tab.~\ref{tab:network-sizes}, we are able to leverage this feature size separation for layer separation tasks such as occlusion, reflection, or shadow removal. 

For tasks such as video segmentation, it is important that both the transmission layer and obstruction layer are able to represent high-resolution images, as the purpose here is to divide and compress video content into two canonical views, alpha matte, and optical flow. Hence for the video segmentation task in Tab.~\ref{tab:network-sizes} both layers have \textit{Large} network encodings. Conversely, for a task such as shadow removal we want to minimize the amount of color and alpha information the shadow obstruction layer is able to represent -- as shadows, like the mask example in Fig.~\ref{fig:supp_image_fitting}, are simple-to-fit image features. Correspondingly, the shadow removal task in Tab.~\ref{tab:network-sizes} has a \textit{Tiny} image color encoding and only a \textit{Medium} size alpha encoding. We keep these \emph{parameters constant between all tested scenes} for clarity of presentation, however we emphasize that these model configurations are not prescriptive; all neural scene fitting approaches~\cite{mildenhall2020nerf} have per-scene optimal parameters. Given the relatively fast training speed of our approach, approximately 10mins on a single Nvidia RTX 4090 GPU to fit a scene, in settings where data acquisition is costly  -- e.g., scientific imaging settings such as microscopy -- it may even be tractable to sweep model parameters to optimally reconstruct each individual capture.

%% file: supp_1_results.tex
\section{Additional Reconstruction Results}
\label{sec:supp_results}
\begin{figure*}[t!]
    \centering
    \includegraphics[width=\linewidth]{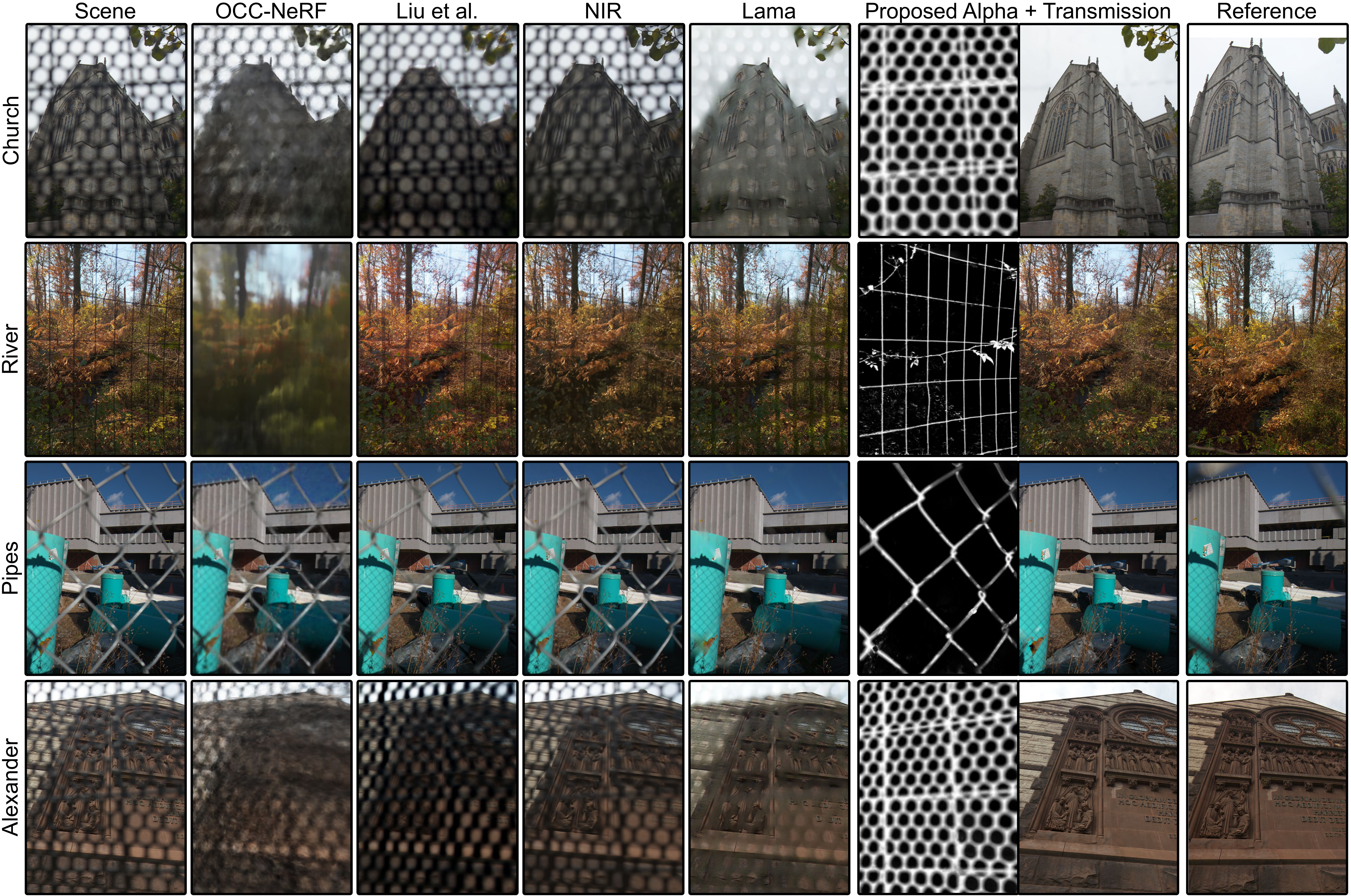}
    \caption{Occlusion removal results and estimated alpha maps for a set of captures with reference views, with comparisons to single image,
multi-view, and NeRF fitting approaches. See video materials for visualization of input data and scene fitting.}
    \label{fig:supp_results_occlusion}
    \vspace{-1.5em}
\end{figure*}

\noindent In this section, we provide additional quantitative and qualitative obstruction removal results, comparing our proposed model against a range of multi-view and single-image methods. We include discussion of challenging imaging settings and potential directions of future work to address them.\\
\noindent\textbf{Occlusion Removal}\hspace{0.1em} Focusing on natural environmental occluders such as fences and grates, we include a set of additional occlusion removal results in Fig.~\ref{fig:supp_results_occlusion}. We evaluate our results against a multi-image learning-based obstruction removal method Liu et al.~\cite{liu2020learning}, a NeRF-based method OCC-NeRF~\cite{zhu2023occlusion}, the flow plus homography neural image representation NIR~\cite{nam2022neural}, and a single image inpainting approach Lama~\cite{suvorov2021resolution} -- to which we provide hand-drawn masks of the occlusion. We find that, as observed in the main text, the multi-image methods struggle to remove significant parts of the obstruction. Though in some scenes, the multi-image baselines are able to decrease the opacity of the occluder to reveal details behind it. Nevertheless, in all cases the obstruction is still clearly visible after applying each baseline. Given the small camera baseline setting of our input data, the volumetric OCC-NeRF approach struggles to converge on a cohesive 3D scene representation, producing blurred or otherwise inconsistent image reconstructions -- as is the case for the \textit{Digger} scene. We find that the the homography-based NIR method also struggles in this small baseline setting, often identifying the entire scene as the canonical view rather than partly obstructed.
Given hand annotated masks, single image methods such as DALL·E and Lama~\cite{suvorov2021resolution} can successfully inpaint sparse occluders such as the fences in the \textit{Digger} and \textit{Pipes}  scenes, but struggle to recover content behind dense occluders such as \textit{Hallway} and \textit{Desk} in Fig.~\ref{fig:supp_results_occlusion}. As they have no way to aggregate content between frames, they ``recover" hidden content from visual priors on the scene, which may not be reliable when the scene is severely occluded.

\begin{figure}[h!]
    \centering
    \includegraphics[width=\linewidth]{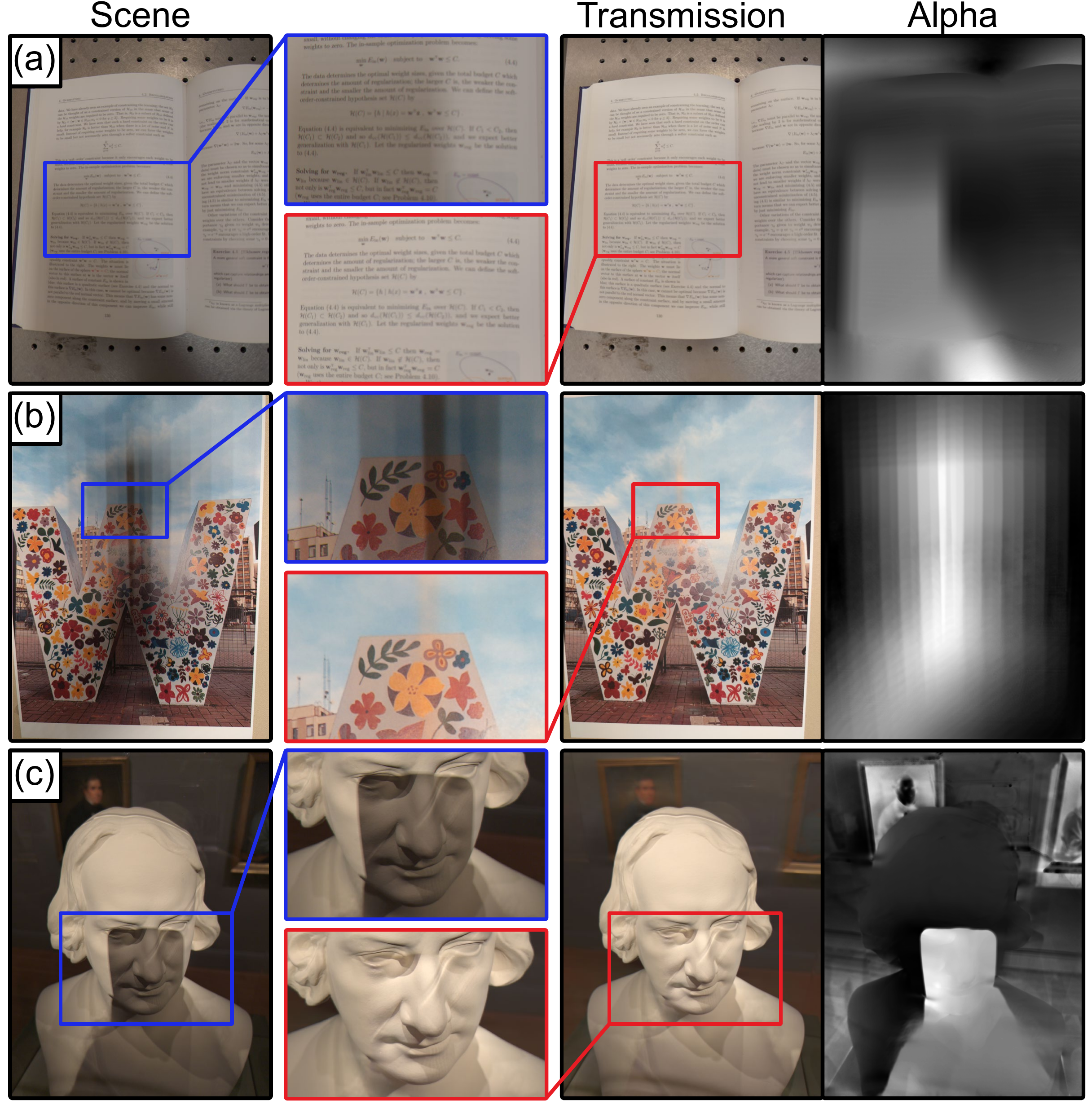}
    \vspace{-1.5em}
    \caption{Shadow removal results under different lighting conditions: (a) partially diffuse, (b) multiple point, (c) single point.}
    \label{fig:supp_shadow}
    \vspace{-1.5em}
\end{figure}

In contrast, our method automatically distills a high-quality alpha matte for the obstruction and reconstructs the underlying transmission layer using information from multiple views. This mask is of similar quality regardless of whether the scene is obstructed by a dense occluder or a sparse occluder, so long as there is sufficient parallax between the two layers. The depth-separation properties of our alpha estimation are showcased in the \textit{River} example, where the obstruction layer isolated not only the grid of the fence, but also the branches and leaves weaved through the fence. Our method reconstructs the transmitted layer behind the occlusion with favorable results compared to all baseline methods.

\begin{figure*}[h!]
    \centering
    \includegraphics[width=\linewidth]{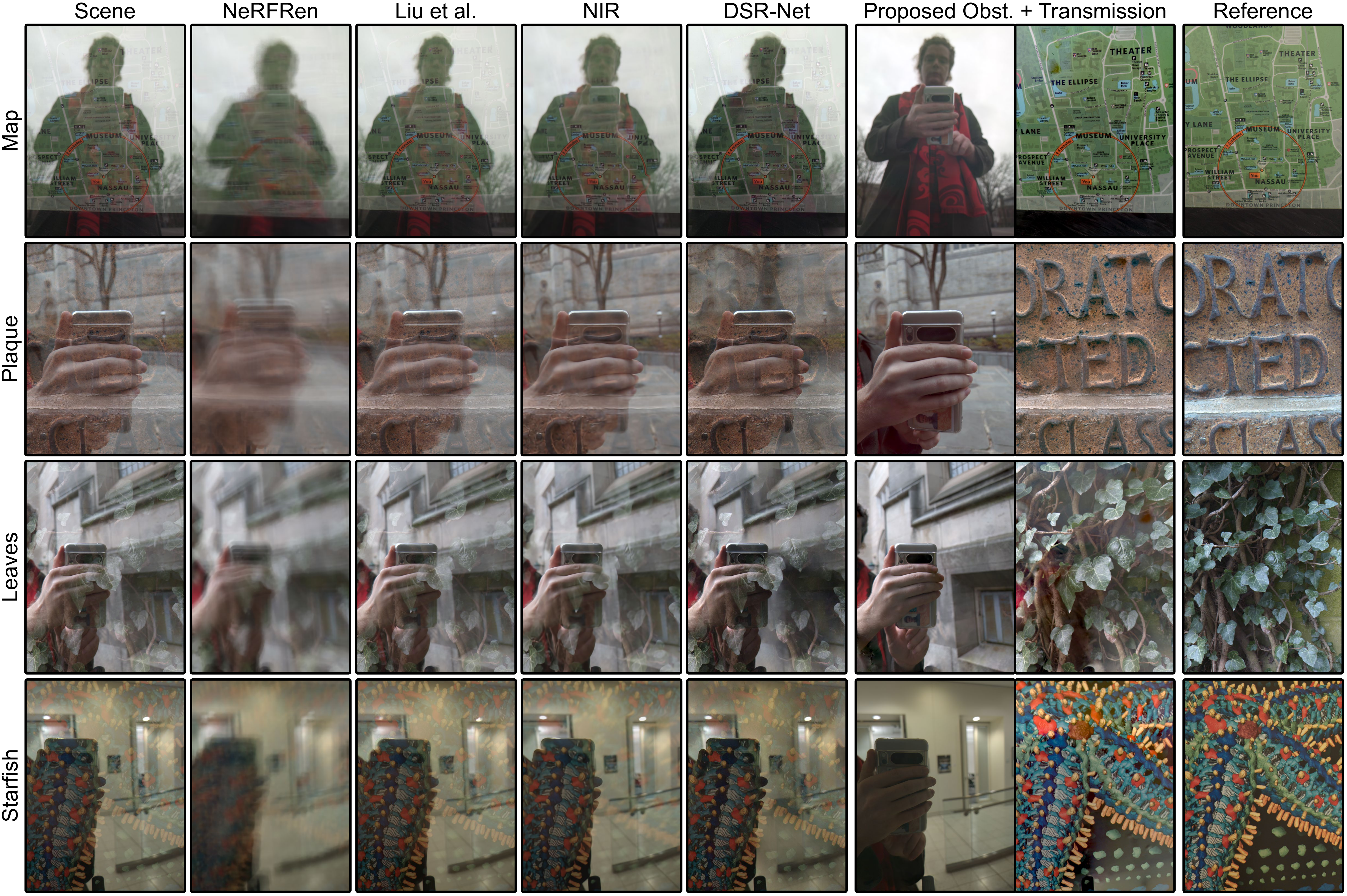}
    \vspace{-1.5em}
    \caption{Reflection removal results and estimated alpha maps for a set of captures with reference views, with comparisons to single image,
multi-view, and NeRF fitting approaches. See video materials for visualization of input data and scene fitting.}
    \label{fig:supp_results_reflection}
    \vspace{-1.5em}
\end{figure*}

\begin{figure}[h!]
    \centering
    \includegraphics[width=\linewidth]{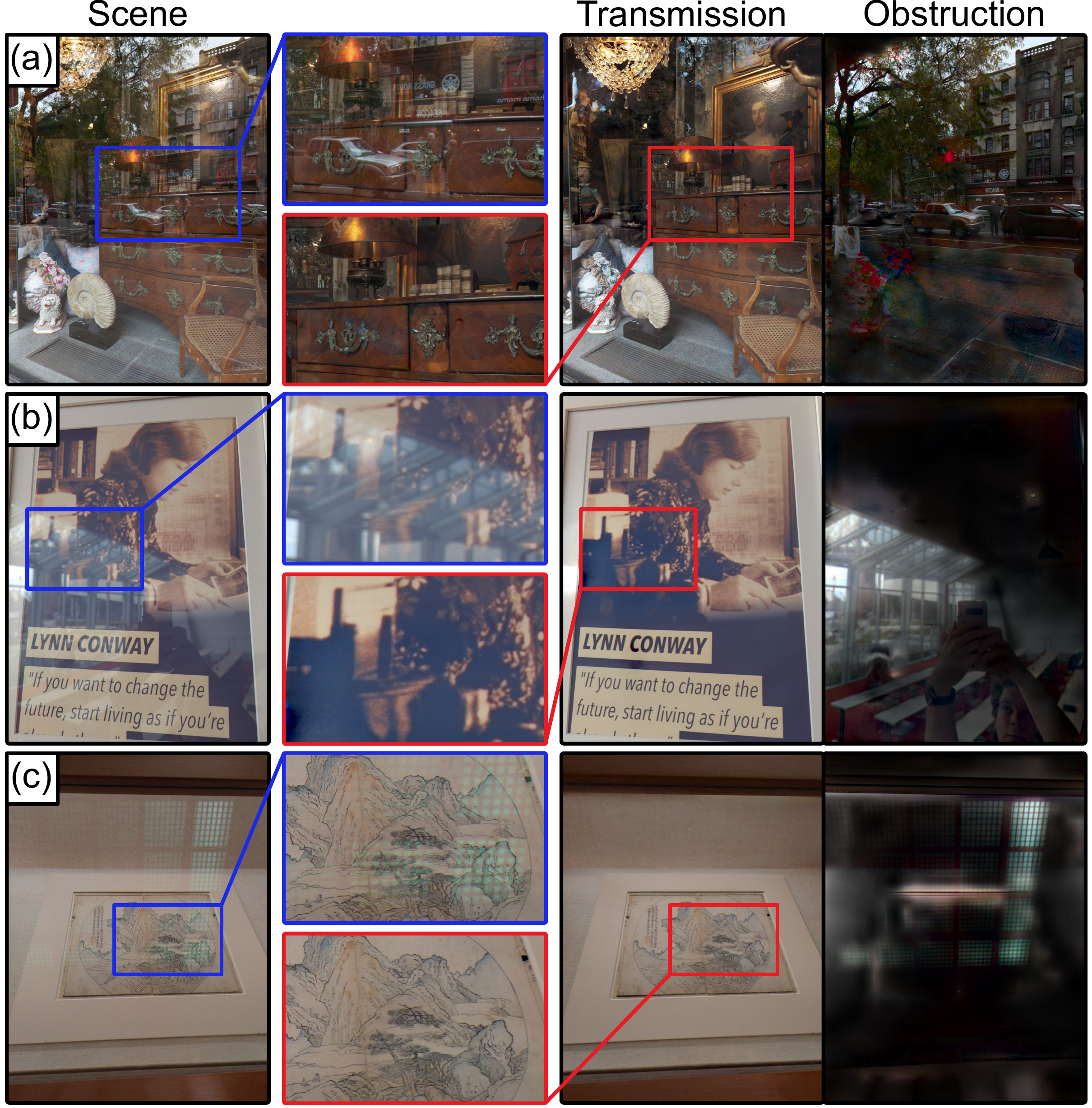}
    \vspace{-1.5em}
    \caption{Reflection removal results for challenging in-the-wild scenes: (a) storefront window, (b) poster, (c) museum painting.}
    \label{fig:wild_reflection}
    \vspace{-1.5em}
\end{figure}

\noindent\textbf{Reflection Removal}\hspace{0.1em} 
For reflection removal, we compare with the reflection-aware NeRF-based method NeRFReN~\cite{guo2022nerfren} in addition to NIR~\cite{nam2022neural}, Liu et al.~\cite{liu2020learning}, and the single-image reflection removal method DSRNet~\cite{hu2023single}. We show reflection removal results in Fig.~\ref{fig:supp_results_reflection}. We observe results with a similar trend to those in the obstruction removal task. The volumetric method NeRFReN struggles to reconstruct a high-fidelity scene representation, as Liu et al. and NIR also struggle with the small baseline of the camera motion. The single-image method DSRNet performs best among the baselines, as it has no priors on image motion. However, without the ability to draw information from multiple views, DSRNet uses learned priors to disambiguate reflected and transmitted content. This appears to fail for high opacity reflections, such as the \textit{Grass} example, and regions like the top left of the \textit{Test Page} scene where the reflection can be mistaken for a real lighting effect. Our method achieves the highest-quality reconstruction and layer separation among all methods tested, across all scenes, with our estimated alpha matte revealing the detailed structure of the scene being reflected. In Fig.~\ref{fig:wild_reflection} we also showcase our model's performance on challenging, in-the-wild scenes where we do not have the ability to acquire reference views. We observe robust reflection removal, matching the reconstruction quality observed for scenes acquired with our tripod setup.

\begin{figure*}[h!]
    \centering
    \includegraphics[width=\linewidth]{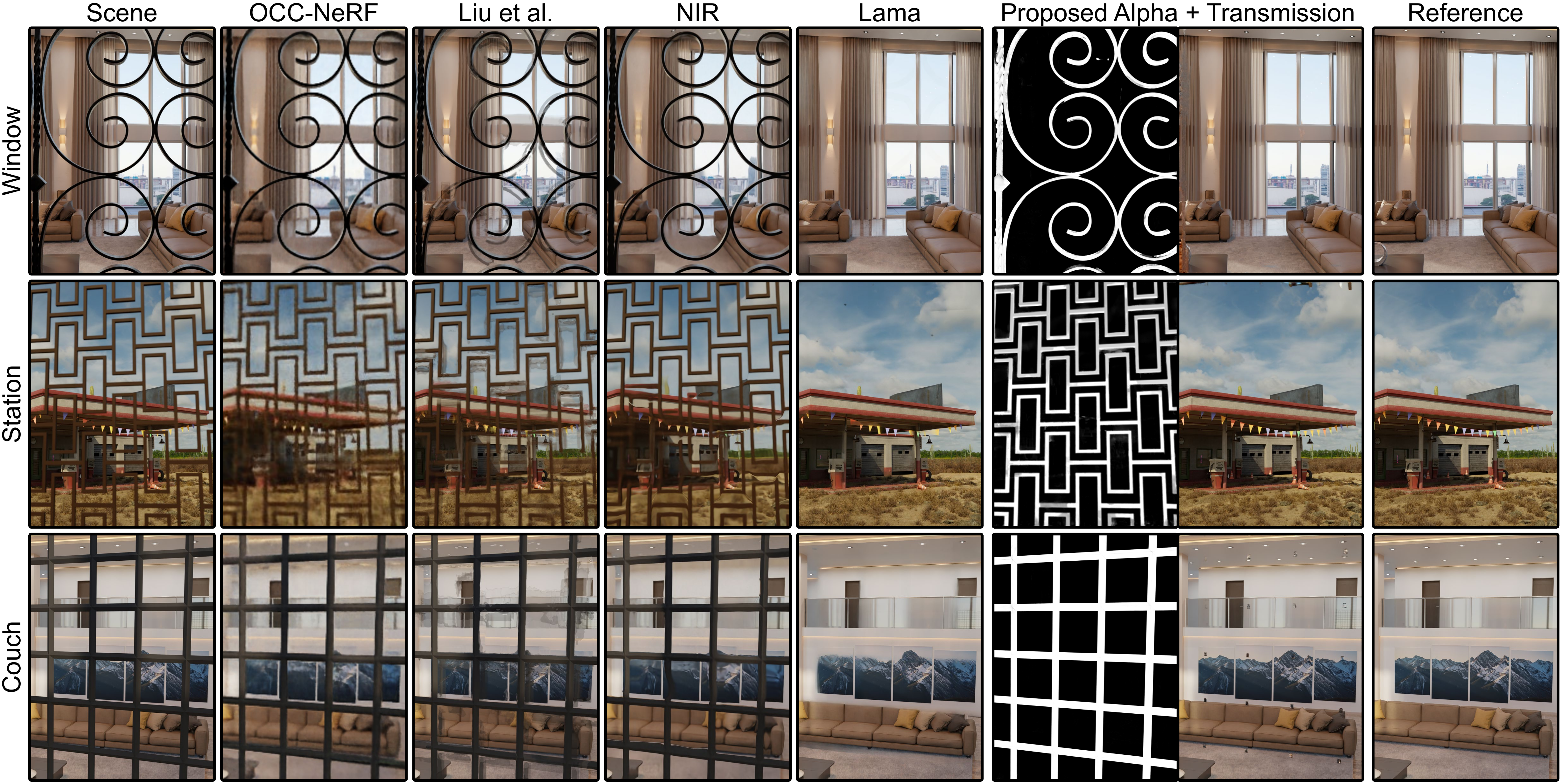}
        \resizebox{\linewidth}{!}{
     \begin{tabular}[b]{lccccclccccc}
     \midrule
          \textbf{Occlusion} & OCC-NeRF & Liu et al. & NIR & Lama & Proposed & \textbf{Occlusion} & OCC-NeRF & Liu et al. & NIR & Lama & Proposed\\
	\midrule
	\midrule
    \textit{Bookstore} & 21.34$/$0.467 & 24.57$/$0.716 & 20.44$/$0.459 & 28.72$/$0.824 & \textbf{35.59/0.929} & \textit{Window}  & 14.22$/$0.608 & 14.03$/$0.718 & 13.98$/$0.680 & 30.44$/$0.939 & \textbf{32.91/0.953}    \\
    \textit{Couch} & 13.37$/$0.658 & 13.09$/$0.708 & 13.09$/$0.690 & 26.86$/$0.927 & \textbf{31.65/0.963} & \textit{Station}   & 14.62$/$0.569 & 14.40$/$0.649 & 14.05$/$0.595 & 26.48$/$0.874 & \textbf{30.25/0.936}     \\
    \bottomrule
  \end{tabular}
  }
    \caption{Qualitative and quantitative occlusion removal results for a set of 3D rendered scenes with paired ground truth. Evaluation metrics formatted as PSNR/SSIM.}
    \label{fig:supp_occlusion_synthetic}
    \vspace{-1em}
\end{figure*}

\begin{figure*}[h!]
    \centering
    \includegraphics[width=\linewidth]{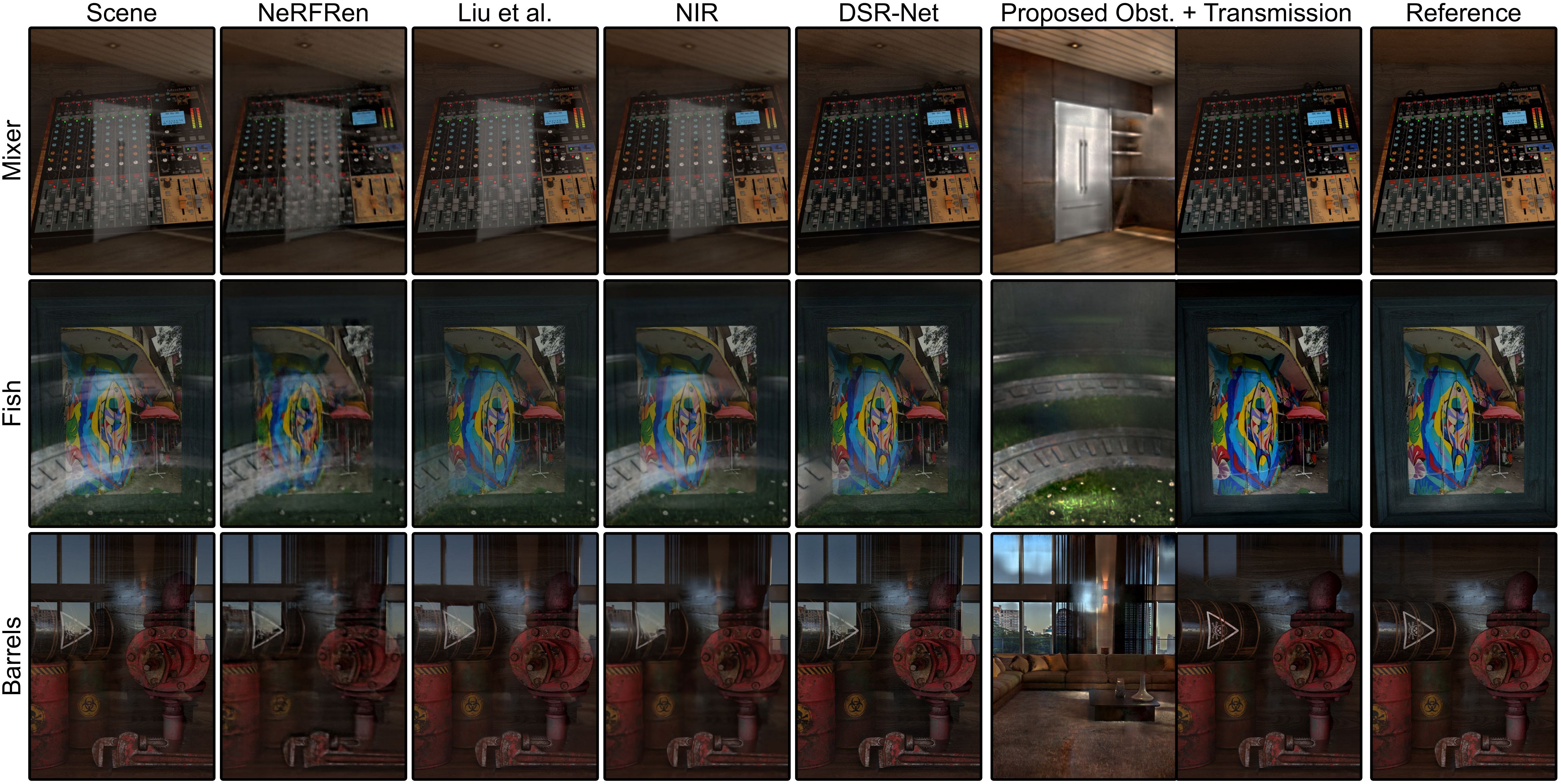}
 \resizebox{\linewidth}{!}{
     \begin{tabular}[b]{lccccclccccc}
     \midrule
          \textbf{Reflection} & NeRFReN & Liu et al. & NIR & DSR-Net & Proposed & \textbf{Reflection} & NeRFReN & Liu et al. & NIR & DSR-Net & Proposed\\
	\midrule
	\midrule
    \textit{Vase} & 18.40$/$0.688 & 18.53$/$0.849 & 18.52$/$0.771 & 19.43$/$0.835 & \textbf{27.20/0.942} & \textit{Barrels}  & 20.59$/$0.736 & 20.76$/$0.850 & 20.57$/$0.752 & 20.39$/$0.811 & \textbf{26.07/0.921}    \\
    \textit{Fish} & 20.05$/$0.744 & 24.32$/$0.904 & 19.78$/$0.783 & 23.51$/$0.875 & \textbf{25.47/0.824} & \textit{Mixer}   & 19.57$/$0.628 & 19.20$/$0.726 & 19.34$/$0.660 & 24.56$/$0.794 & \textbf{27.06/0.875}     \\
    \bottomrule
  \end{tabular}
  }
\vspace*{-1.5em}
    \caption{Qualitative and quantitative reflection removal results for a set of 3D rendered scenes with paired ground truth. Evaluation metrics formatted as PSNR/SSIM.}
    \label{fig:supp_reflection_synthetic}
    \vspace{-1em}

\end{figure*}

\noindent\textbf{Validation on Synthetic Scenes}\hspace{0.1em}
Next, we discuss evaluations on synthetic data with known ground truth. As described in Sec.~\ref{sec:supp_implementation}, we compare our obstruction and reflection removal method with several baselines, including OCC-NeRF~\cite{zhu2023occlusion}, Liu et al.~\cite{liu2020learning}, and NIR~\cite{nam2022neural}, in Fig.~\ref{fig:supp_occlusion_synthetic} and Fig.~\ref{fig:supp_reflection_synthetic} respectively. We provide NeRF-based methods with ground truth camera poses, which results in a higher fidelity NeRF-based reconstruction than on real-world data. We also include quantitative metrics measuring the quality of the reconstructed transmitted layer. In our evaluations of obstruction removal, we observe that just as in the real-world examples, the multi-image based methods almost entirely fail to remove occlusions. Single-image methods, when provided with a ground-truth occlusion mask, are able to construct a coherent transmission layer. However, upon closer inspection the single-image results are missing details in the ground-truth transmission layer, such as the string of flags in \textit{Station}, in Fig.~\ref{fig:supp_occlusion_synthetic}. Our reconstructions have the highest PSNR and SSIM across all methods and all scenes tested, both removing the obstruction and accurately recovering details hidden behind it. We also observe that most multi-image methods fail to remove reflections in Fig.~\ref{fig:supp_reflection_synthetic}, with the exception of Liu et al. on the scene \textit{Vase Outdoor}. The single-image method DSRNet~\cite{hu2023single} once again outperforms the multi-view baselines, removing the reflection in front of the rightmost vase in \textit{Vase Outdoor}, and improving the visibility into the glass case in \textit{Barrels}. However, similarly to the real-world scenes, DSRNet struggles to fully disambiguate more opaque reflections like the cabinet doors in \textit{Mixer}. Our proposed approach is able the only method that removes the reflection in front of the leftmost vase in \textit{Vase Indoor}, and the entire reflection in \textit{Mixer}. This qualitative performance is reflected in the quantitative metrics, where we achieve the highest PSNR and SSIM for all tested scenes.\\
\noindent\textbf{Shadow Removal}\hspace{0.1em} In Fig.~\ref{fig:supp_shadow} we demonstrate shadow removal results for scenes with disparate lighting conditions: (a) a book illuminated by a diffuse overhead lamp, (b) a poster illuminated by an array of LEDs, and (c) a bust illuminated by a strong point light source. We note that the grid of LEDs act as a set of point light sources, producing multiple copies of the shadow to be overlayed on the scene. In all settings we are able to extract the shadow with the same obstruction network defined in the \textit{shadow removal} application in Tab.~\ref{tab:application-configs}, further reinforcing the our image fitting findings from Fig.~\ref{fig:supp_image_fitting}. Namely that coordinate networks with low-resolution multi-resolution hash encodings are able to effectively fit both scenes comprised of smooth gradients, as in the diffuse shadow case, and limited numbers of image discontinuities, as in the multiple point source case. In (c) we furthermore see that while the photographer-cast shadow is successfully removed from the bust, the shadows cast by other light sources are left intact. This reinforces that our proposed model is separating shadows based not only on their color, but on the motion they exhibit in the scene; as the other shadows cast on the bust undergo the same parallax motion as the bust itself.

\begin{figure}[h!]
    \centering
    \includegraphics[width=\linewidth]{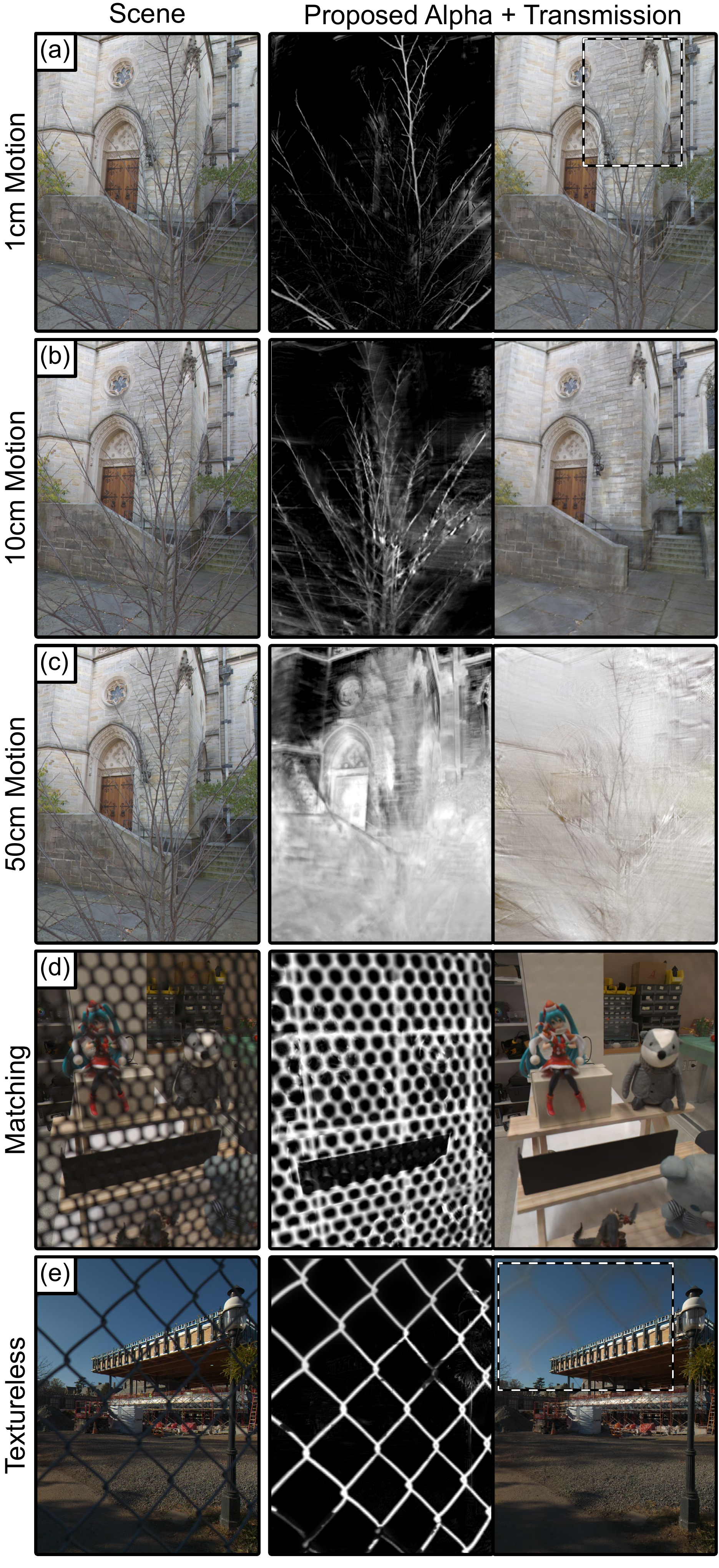}
    \caption{Challenging image reconstruction cases including varying scales of camera motion, overlap between occluder and transmission colors, and residual signal left on scene content in low-texture regions. Areas of interest highlighted with dashed border.}
    \label{fig:supp_challenging}
    \vspace{-1.5em}
\end{figure}

\noindent\textbf{Challenging Settings}\hspace{0.1em} We compile a set of challenging imaging settings in Fig.~\ref{fig:supp_challenging} which highlight areas where our proposed approach could be improved. One limitation of our work is that it cannot generate unseen content. While this means it cannot hallucinate content from unreliable image priors, it also means that it is highly parallax-dependent for generating accurate reconstructions. This is highlighted in Fig.~\ref{fig:supp_challenging} (a-c), where with hand motion on the scale of 1cm is only enough to separate and remove the topmost branch of the occluding plant. Motion on the scale of 10cm is enough to remove most of the branches, but larger motion on the scale of half a meter in diameter causes the reconstruction to break down. This is likely due to the small motion and angle assumptions in our camera model, as it is not able to successfully jointly align the input image data and learn its multi-layer representation. Thus work on large motion or wide-angle data for large obstruction removal --  e.g., removing telephone poles from in from  -- remains an open problem. Fig.~\ref{fig:supp_challenging} (d) demonstrates the challenge of estimating an accurate alpha matte when the transmitted and obstructed content are matching colors. In this case, although the obstruction is ``removed", we see that the alpha matte is missing a gap around the black object in the scene behind the occluder. In this region the model does not need to use the obstruction layer to represent pixels that are already black in the transmission layer -- in fact, the alpha regularization term $\mathcal{R}_\alpha$ would penalize this. Thus the alpha matte is actually a produce of both the actual alpha of the obstruction and its relative color difference with what it is occluding. Fig.~\ref{fig:supp_challenging} (e) highlights a related problem. In regions where the transmission layer is low-texture, and lacks parallax cues, it is ambiguous what is being obstructed and where the border of the obstruction lies. Thus ghosting artifacts are left behind in areas such as the sky of the \textit{Textureless} scene. What is noteworthy, however, is that these are also exactly the regions in which in-painting methods such as Lama~\cite{suvorov2021resolution} are most successful, as there are no complex textures that need to be recovered from incomplete data, leaving a hybrid model as an interesting direction for future work.


\noindent

%% file: supp_2_experiments.tex
\section{Additional Experiments and Analysis}
\label{sec:supp_experiments}
\begin{figure*}[t]
    \centering
    \includegraphics[width=\linewidth]{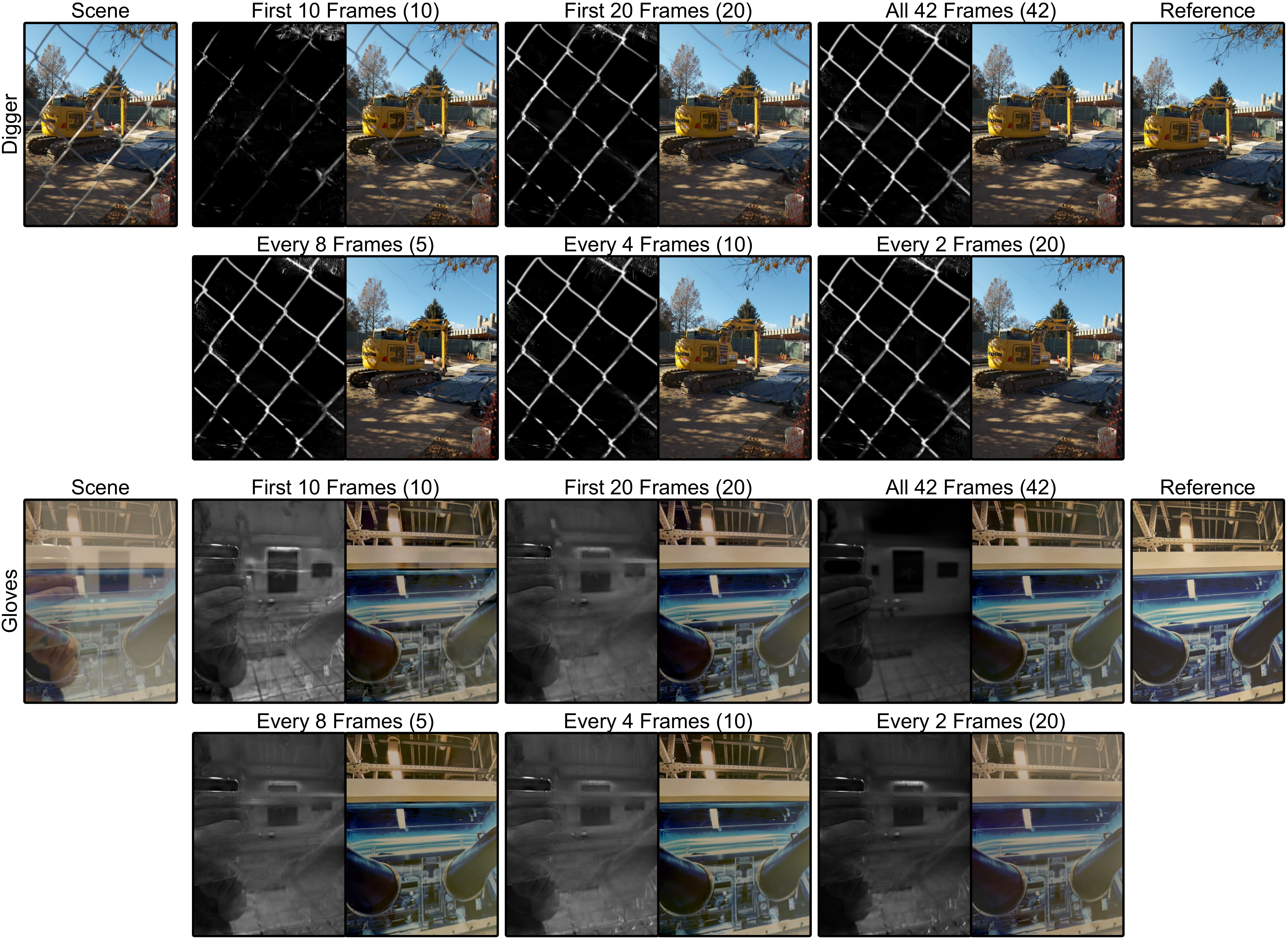}
    \vspace{-1.5em}
    \caption{Ablation study on the effects of the number of input frames or duration of capture on transmission layer reconstruction and estimated alpha matte. Total number of frames input into the model denoted by the number in parentheses-- e.g., (10) = ten frames.}
    \label{fig:supp_ablation_frames}
\vspace{-0.5em}
\end{figure*}
\begin{figure*}[h!]
    \centering
    \includegraphics[width=\linewidth]{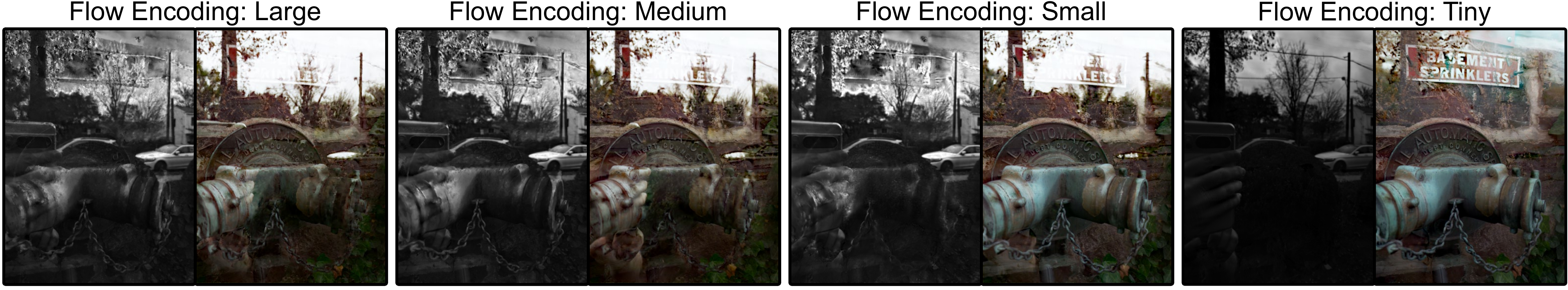}
    \vspace{-1.5em}
    \caption{Ablation study on the effects of flow encoding size (Tab.~\ref{tab:network-sizes}) on transmission layer reconstruction and estimated alpha matte.}
    \vspace{-1.5em}
    \label{fig:supp_ablation_network}
\end{figure*}

\begin{figure*}[t]
    \centering
    \includegraphics[width=\linewidth]{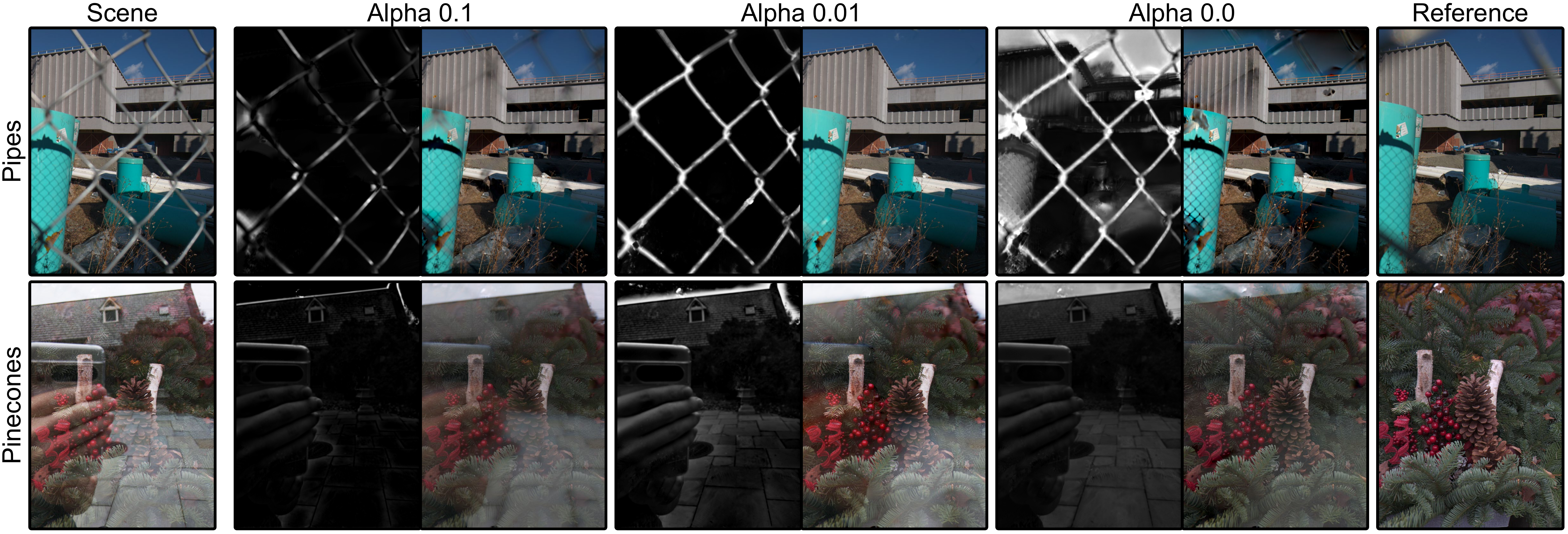}
    \caption{Ablation study on the effects of alpha regularization weight $\eta_\alpha$ on transmission layer reconstruction and estimated alpha matte.}
    \label{fig:supp_ablation_alpha}
    \vspace{-2em}
\end{figure*}
\begin{figure}[h!]
    \centering
    \includegraphics[width=\linewidth]{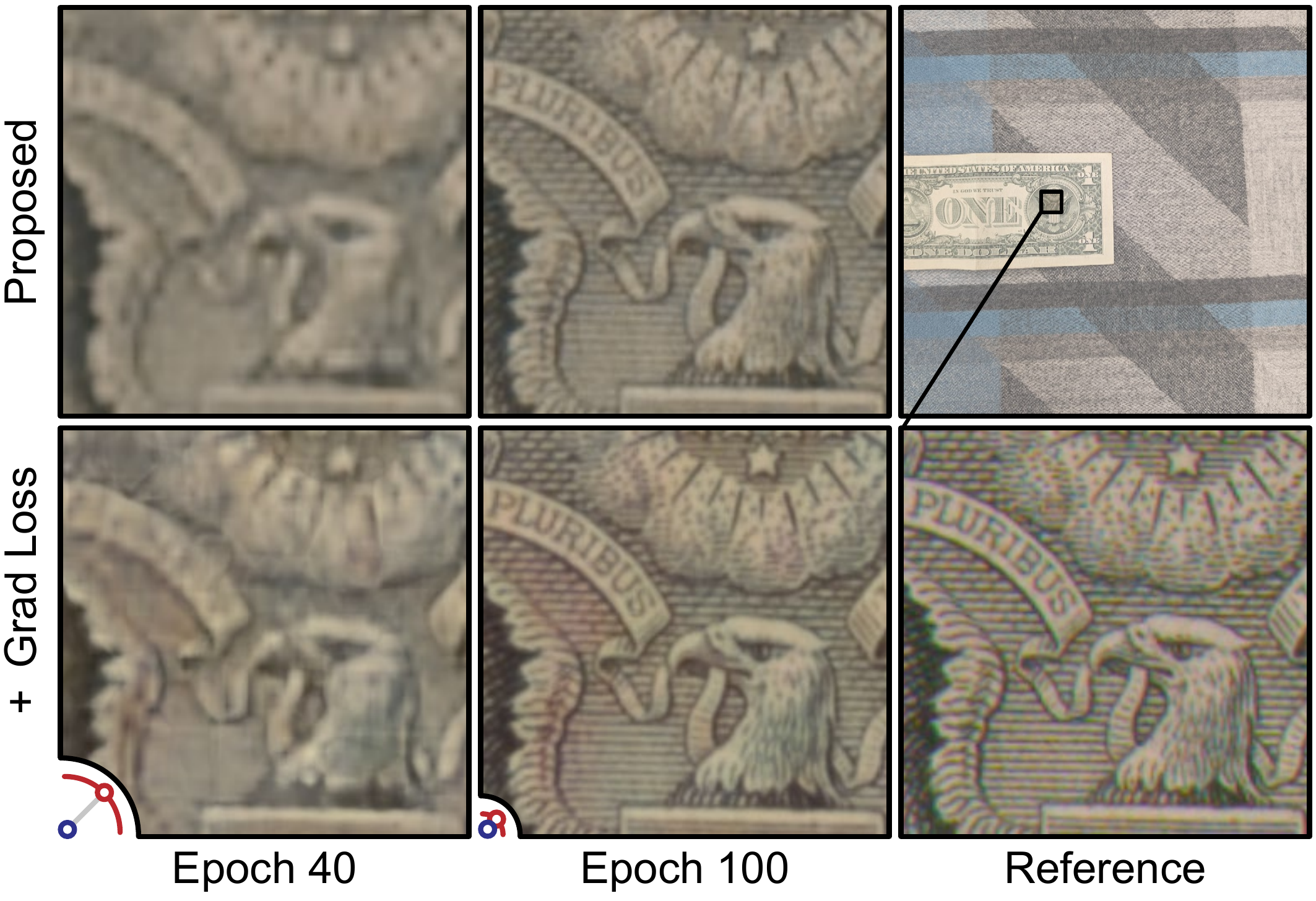}
    \caption{Visualization of the effects of gradient loss $\mathcal{L}_\textsc{g}$ on image reconstruction at 25x zoom. Inset bottom left is the radius of perturbation at epoch 40 and epoch 100, the end of training. }
    \label{fig:supp_gradient_loss}
    \vspace{-1.5em}
\end{figure}
\noindent\textbf{Gradient Loss}\hspace{0.1em} A significant challenge posed by the task of aggregating long-burst data is the so-called problem of ``regression to the mean''. When minimizing a metric such as relative mean-square error, which penalizes small color differences significantly less than large discrepancies, the final reconstruction is encouraged to be smoother than the original input data~\cite{bahat2020explorable}. Thus, in developing our approach we explored -- but ultimately did not use -- a form of gradient penalty loss:
\begin{align}
    \mathcal{L}_\textsc{g} &= |(\scalebox{0.7}{$\Delta$}c - \scalebox{0.7}{$\Delta$}\hat c)/(\mathrm{sg}(\scalebox{0.7}{$\Delta$}c) + \epsilon)|^2 \nonumber.
\end{align}
Rather than sample a grid of points around $u^\textsc{o},v^\textsc{o}$ and $u^\textsc{t},v^\textsc{t}$ or perform a second pass over the image networks~\cite{nam2022neural} to compute Jacobians, we compute color gradients $\scalebox{0.7}{$\Delta$} c$ by pairing each ray with an input perturbed in a random direction
\begin{align}\label{eq:perturb}
    \scalebox{0.7}{$\Delta$} c &= I(u,v,t) - I(\tilde u,\tilde v,t) \\
    \tilde u,\tilde v &= u + r\mathrm{cos}(\phi), \,\, v + r\mathrm{sin}(\phi), \quad \phi \sim \mathcal{U}(0,2\pi)\nonumber,
\end{align}
where $r$ determines the magnitude of the perturbation. The estimated color gradient $\scalebox{0.7}{$\Delta$} \hat c$ is similarly calculated for the output colors of our model. Illustrated in Fig.~\ref{fig:supp_gradient_loss}, by reducing radius $r$ from multi-pixel to sub-pixel perturbations during training, we are able to improve fine feature recovery in the final reconstruction via gradient loss $\mathcal{L}_\textsc{g}$ without significantly impacting training time -- as perturbed samples are also re-used for regular photometric loss calculation $\mathcal{L}_p$. However, as we do not apply any demosaicing or post-processing to our input Bayer array data, we find this loss can also lead to increased color-fringing artifacts -- the red tint in the bottom row of Fig.~\ref{fig:supp_gradient_loss}. For these reasons, and poor convergence in noisy scenes, we did not include this loss in the final model. However, there may be potentially interesting avenue of future research into a jointly trained demosaicing module to robustly estimate real color gradient directly from quantized and discretized Bayer array values. 



\noindent\textbf{Alpha Regularization Ablation}\hspace{0.1em} In Fig.~\ref{fig:supp_ablation_alpha}, we visualize the effects of alpha regularization weight $\eta_\alpha$ on reconstruction. The primary function of this regularization is remove low-parallax content from the obstruction layer, as there is no alpha penalty for reconstructing the same content via the transmission layer. As seen in the \textit{Pipes} example, without alpha regularization the obstruction layer is able to freely reconstruct part of the transmitted scene content such as the sky, the pipes, and the walls of the occluded buildings. A small penalty of $\eta_\alpha=0.01$ is enough to remove this unwanted content from the obstruction layer, while $\eta_\alpha=0.1$ is enough to also start removing parts of the actual obstruction. Contrastingly, in the case of reflection scenes such as \textit{Pinecones}, even a relatively small alpha regularization weight of $\eta_\alpha=0.01$ removes part of the actual reflection -- leaving behind a grey smudge in the bottom right corner of the reconstruction. As reflections are typically partially transparent obstructions, and can occupy a large area of the scene, removing them purely photometrically is ill-conditioned. There is no visual difference between a gray reflector covering the entire view of the camera and the scene actually being gray. Thus $\eta_\alpha$ can also be a user-dependent parameter tuned to the desired ``amount'' of reflection removal.\\
\noindent\textbf{Frame Count Ablation}\hspace{0.1em} Thusfar we have used all 42 frames in each long-burst capture as input to our method, but we highlight that this is not a requirement of the approach. The training process can be applied to any number of frames -- within computational limits. In Fig.~\ref{fig:supp_ablation_frames} we showcase reconstruction results for both subsampled captures, where only every $k$-th frame of the image sequence is kept for training, and shortened captures, where only the first $n$ frames are retained. Similar to the problem of depth reconstruction~\cite{chugunov2023shakes}, we find that obstruction removal performance directly depends on the total amount of parallax in the input. Sampling the \textit{first} 10 frames -- approximately 0.5 seconds of recording -- results in diminished obstruction removal for both the \textit{Digger} and \textit{Gloves} scenes as the obstruction exhibits significantly less motion during the capture. In contrast, given \textit{a five frame input sampled evenly across the full two-second capture}, our proposed approach is able to successfully reconstruct and remove the obstruction. This subsampled scene also \textit{trains considerably faster}, converging in only 3 minutes as less frames need to be sampled per batch -- or equivalently more rays can be sampled from each frame for each iteration. This further validates the benefit of a long burst capture.\\
\begin{figure}[t]
    \centering
    \includegraphics[width=\linewidth]{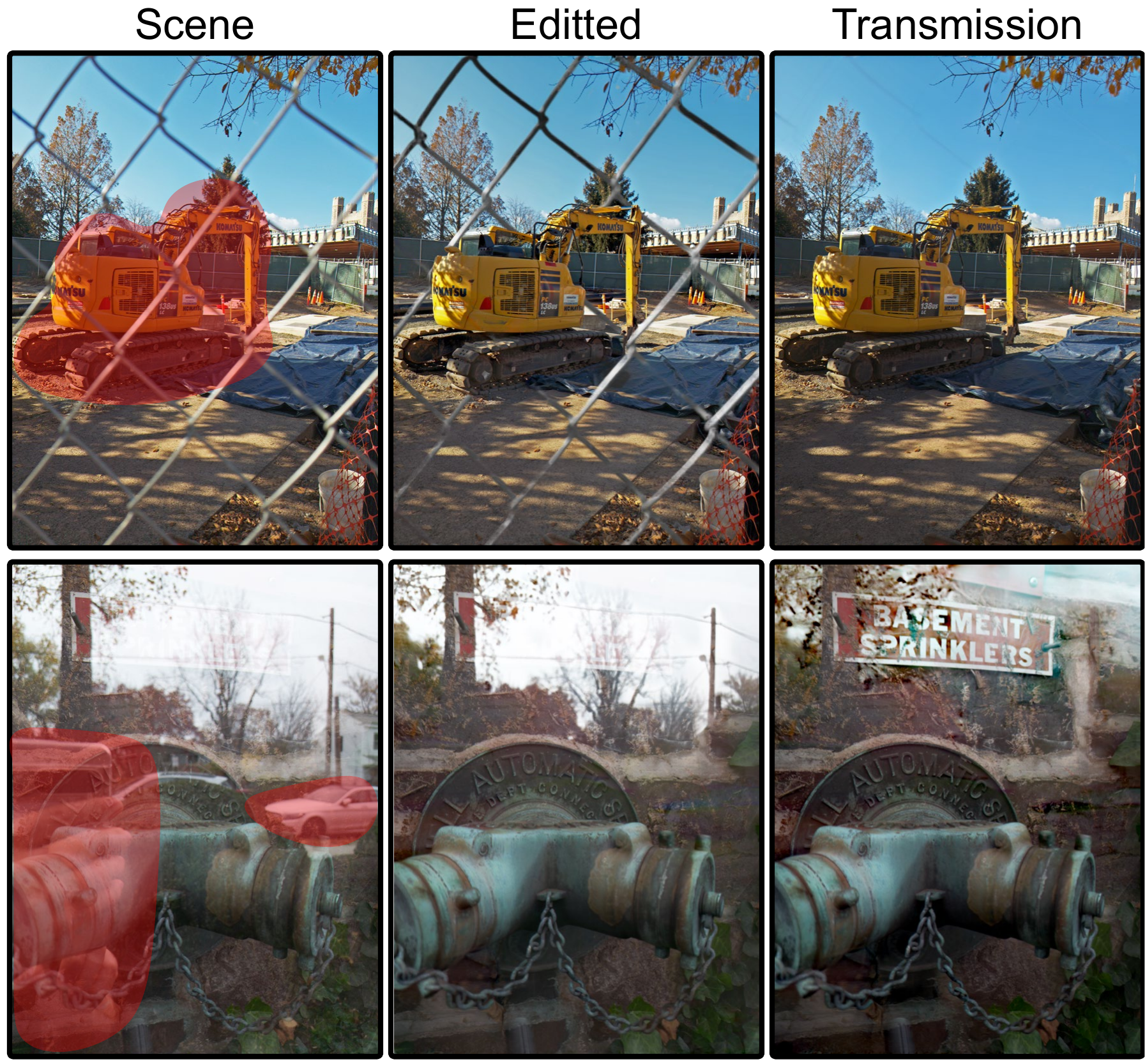}
    \caption{Demonstration of user-interactive scene editing facilitated by layer separation. Only the user-selected region of the obstruction, highlighted in red, is removed without affecting surrounding scene content, see text.}
    \label{fig:supp_editting}
    \vspace{-1.5em}
\end{figure}
\noindent\textbf{Flow Encoding Size}\hspace{0.1em} A key model parameter which controls layer separation, as discussed in Section ~\ref{sec:supp_implementation}, is the size of the encoding for our neural spline flow fields. In Fig.~\ref{fig:supp_ablation_network} we illustrate the effects on obstruction removal of over-parameterizing this flow representation. When the two layers are undergoing simple motion caused by parallax from natural hand tremor, a \textit{Tiny} flow encoding is able to represent and pull apart the motion of the reflected and transmitted content. However, high-resolution neural spline fields, just like a traditional flow volume $h(u,v,t)$, can quickly overfit the scene and mix content between layers. We can see this clearly in the \textit{Large} flow encoding example where the reflected phone, trees, and parked car appear in both the obstruction alpha matte and transmission image. Thus it is critical to the success of our method to construct a task-specific neural spline field representation appropriate for the expected amount and density of scene motion.

\noindent\textbf{Applications to Scene Editing}\hspace{0.1em} In Fig.~\ref{fig:supp_editting} we showcase the scene editing functionality facilitated by our proposed methods layer separation. As we estimate an image model for both the transmission and obstruction, we are not limited to only removing a layer but can independently manipulate them. In this example we rasterize both layers to RGBA images and input them into an image editor. The user is then able to highlight and delete a portion of the occlusion while retaining its other content. Thus we can create physically unrealizable photographs such as only the fence appearing to be behind the \textit{Digger}, or selectively remove the photographer's hand and parked car from the \textit{Hydrant} scene.

%% file: cvpr.bbl
\begin{thebibliography}{10}\itemsep=-1pt

\bibitem{adeel2022defencing}
Hannan Adeel, Muhammad~Mohsin Riaz, and Syed~Sohaib Ali.
\newblock De-fencing and multi-focus fusion using markov random field and image inpainting.
\newblock {\em IEEE Access}, 10:35992--36005, 2022.

\bibitem{bahat2020explorable}
Yuval Bahat and Tomer Michaeli.
\newblock Explorable super resolution.
\newblock In {\em Proceedings of the IEEE/CVF Conference on Computer Vision and Pattern Recognition}, pages 2716--2725, 2020.

\bibitem{barron2021mip}
Jonathan~T Barron, Ben Mildenhall, Matthew Tancik, Peter Hedman, Ricardo Martin-Brualla, and Pratul~P Srinivasan.
\newblock Mip-nerf: A multiscale representation for anti-aliasing neural radiance fields.
\newblock In {\em Proceedings of the IEEE/CVF International Conference on Computer Vision}, pages 5855--5864, 2021.

\bibitem{barron2023zip}
Jonathan~T Barron, Ben Mildenhall, Dor Verbin, Pratul~P Srinivasan, and Peter Hedman.
\newblock Zip-nerf: Anti-aliased grid-based neural radiance fields.
\newblock {\em arXiv preprint arXiv:2304.06706}, 2023.

\bibitem{bertero2021introduction}
Mario Bertero, Patrizia Boccacci, and Christine De~Mol.
\newblock {\em Introduction to inverse problems in imaging}.
\newblock CRC press, 2021.

\bibitem{bhat2021deep}
Goutam Bhat, Martin Danelljan, Luc Van~Gool, and Radu Timofte.
\newblock Deep burst super-resolution.
\newblock In {\em Proceedings of the IEEE/CVF Conference on Computer Vision and Pattern Recognition}, pages 9209--9218, 2021.

\bibitem{blahnik2021smartphone}
Vladan Blahnik and Oliver Schindelbeck.
\newblock Smartphone imaging technology and its applications.
\newblock {\em Advanced Optical Technologies}, 10(3):145--232, 2021.

\bibitem{chen2013simple}
Qifeng Chen and Vladlen Koltun.
\newblock A simple model for intrinsic image decomposition with depth cues.
\newblock In {\em Proceedings of the IEEE international conference on computer vision}, pages 241--248, 2013.

\bibitem{chugunov2023shakes}
Ilya Chugunov, Yuxuan Zhang, and Felix Heide.
\newblock Shakes on a plane: Unsupervised depth estimation from unstabilized photography.
\newblock In {\em Proceedings of the IEEE/CVF Conference on Computer Vision and Pattern Recognition}, pages 13240--13251, 2023.

\bibitem{chugunov2022implicit}
Ilya Chugunov, Yuxuan Zhang, Zhihao Xia, Xuaner Zhang, Jiawen Chen, and Felix Heide.
\newblock The implicit values of a good hand shake: Handheld multi-frame neural depth refinement.
\newblock In {\em Proceedings of the IEEE/CVF Conference on Computer Vision and Pattern Recognition}, pages 2852--2862, 2022.

\bibitem{delbracio2021mobile}
Mauricio Delbracio, Damien Kelly, Michael~S Brown, and Peyman Milanfar.
\newblock Mobile computational photography: A tour.
\newblock {\em arXiv preprint arXiv:2102.09000}, 2021.

\bibitem{farid2016image}
Muhammad~Shahid Farid, Arif Mahmood, and Marco Grangetto.
\newblock Image de-fencing framework with hybrid inpainting algorithm.
\newblock {\em Signal, Image and Video Processing}, 10:1193--1201, 2016.

\bibitem{farin2002curves}
Gerald~E Farin.
\newblock {\em Curves and surfaces for CAGD: a practical guide}.
\newblock Morgan Kaufmann, 2002.

\bibitem{gallo2015locally}
Orazio Gallo, Alejandro Troccoli, Jun Hu, Kari Pulli, and Jan Kautz.
\newblock Locally non-rigid registration for mobile hdr photography.
\newblock In {\em Proceedings of the IEEE conference on computer vision and pattern recognition Workshops}, pages 49--56, 2015.

\bibitem{gandelsman2019double}
Yosef Gandelsman, Assaf Shocher, and Michal Irani.
\newblock " double-dip": unsupervised image decomposition via coupled deep-image-priors.
\newblock In {\em Proceedings of the IEEE/CVF Conference on Computer Vision and Pattern Recognition}, pages 11026--11035, 2019.

\bibitem{godard2018deep}
Cl{\'e}ment Godard, Kevin Matzen, and Matt Uyttendaele.
\newblock Deep burst denoising.
\newblock In {\em Proceedings of the European conference on computer vision (ECCV)}, pages 538--554, 2018.

\bibitem{google2018night}
Google.
\newblock See in the dark with night sight.
\newblock \url{https://blog.google/products/pixel/see-light-night-sight/}, 2018.
\newblock Accessed: 2023-10-24.

\bibitem{google2019astrophotography}
Google.
\newblock Astrophotography with night sight on pixel phones.
\newblock \url{https://blog.research.google/2019/11/astrophotography-with-night-sight-on.html}, 2019.
\newblock Accessed: 2023-10-24.

\bibitem{guo2022nerfren}
Yuan-Chen Guo, Di Kang, Linchao Bao, Yu He, and Song-Hai Zhang.
\newblock Nerfren: Neural radiance fields with reflections.
\newblock In {\em Proceedings of the IEEE/CVF Conference on Computer Vision and Pattern Recognition (CVPR)}, pages 18409--18418, June 2022.

\bibitem{gupta2019fully}
Divyanshu Gupta, Shorya Jain, Utkarsh Tripathi, Pratik Chattopadhyay, and Lipo Wang.
\newblock Fully automated image de-fencing using conditional generative adversarial networks, 2019.

\bibitem{ha2016high}
Hyowon Ha, Sunghoon Im, Jaesik Park, Hae-Gon Jeon, and In~So Kweon.
\newblock High-quality depth from uncalibrated small motion clip.
\newblock In {\em Proceedings of the IEEE conference on computer vision and pattern Recognition}, pages 5413--5421, 2016.

\bibitem{hartley2003multiple}
Richard Hartley and Andrew Zisserman.
\newblock {\em Multiple view geometry in computer vision}.
\newblock Cambridge university press, 2003.

\bibitem{hasinoff2016burst}
Samuel~W Hasinoff, Dillon Sharlet, Ryan Geiss, Andrew Adams, Jonathan~T Barron, Florian Kainz, Jiawen Chen, and Marc Levoy.
\newblock Burst photography for high dynamic range and low-light imaging on mobile cameras.
\newblock {\em ACM Transactions on Graphics (ToG)}, 35(6):1--12, 2016.

\bibitem{hornik1989multilayer}
Kurt Hornik, Maxwell Stinchcombe, and Halbert White.
\newblock Multilayer feedforward networks are universal approximators.
\newblock {\em Neural networks}, 2(5):359--366, 1989.

\bibitem{hu2023single}
Qiming Hu and Xiaojie Guo.
\newblock Single image reflection separation via component synergy.
\newblock In {\em Proceedings of the IEEE/CVF International Conference on Computer Vision (ICCV)}, pages 13138--13147, October 2023.

\bibitem{im2015high}
Sunghoon Im, Hyowon Ha, Gyeongmin Choe, Hae-Gon Jeon, Kyungdon Joo, and In~So Kweon.
\newblock High quality structure from small motion for rolling shutter cameras.
\newblock In {\em Proceedings of the IEEE International Conference on Computer Vision}, pages 837--845, 2015.

\bibitem{kalantari2017deep}
Nima~Khademi Kalantari, Ravi Ramamoorthi, et~al.
\newblock Deep high dynamic range imaging of dynamic scenes.
\newblock {\em ACM Trans. Graph.}, 36(4):144--1, 2017.

\bibitem{kasten2021layered}
Yoni Kasten, Dolev Ofri, Oliver Wang, and Tali Dekel.
\newblock Layered neural atlases for consistent video editing, 2021.

\bibitem{kingma2014adam}
Diederik~P Kingma and Jimmy Ba.
\newblock Adam: A method for stochastic optimization.
\newblock {\em arXiv preprint arXiv:1412.6980}, 2014.

\bibitem{kjolstad2017tensor}
Fredrik Kjolstad, Shoaib Kamil, Stephen Chou, David Lugato, and Saman Amarasinghe.
\newblock The tensor algebra compiler.
\newblock {\em Proceedings of the ACM on Programming Languages}, 1(OOPSLA):1--29, 2017.

\bibitem{kume2023singlefft}
Keitaro Kume and Masaaki Ikehara.
\newblock Single image fence removal using fast fourier transform.
\newblock In {\em 2023 IEEE International Conference on Consumer Electronics (ICCE)}, pages 1--5, 2023.

\bibitem{lecouat2022high}
Bruno Lecouat, Thomas Eboli, Jean Ponce, and Julien Mairal.
\newblock High dynamic range and super-resolution from raw image bursts.
\newblock {\em arXiv preprint arXiv:2207.14671}, 2022.

\bibitem{lei2021robust}
Chenyang Lei and Qifeng Chen.
\newblock Robust reflection removal with reflection-free flash-only cues.
\newblock In {\em IEEE/CVF Conference on Computer Vision and Pattern Recognition (CVPR)}, 2021.

\bibitem{lei2020polarized}
Chenyang Lei, Xuhua Huang, Mengdi Zhang, Qiong Yan, Wenxiu Sun, and Qifeng Chen.
\newblock Polarized reflection removal with perfect alignment in the wild.
\newblock In {\em Proceedings of the IEEE/CVF conference on computer vision and pattern recognition}, pages 1750--1758, 2020.

\bibitem{lei2022robustwild}
Chenyang Lei, Xudong Jiang, and Qifeng Chen.
\newblock Robust reflection removal with flash-only cues in the wild, 2022.

\bibitem{li2013exploiting}
Yu Li and Michael~S. Brown.
\newblock Exploiting reflection change for automatic reflection removal.
\newblock In {\em Proceedings of the IEEE International Conference on Computer Vision (ICCV)}, December 2013.

\bibitem{li2023neuralangelo}
Zhaoshuo Li, Thomas M{\"u}ller, Alex Evans, Russell~H Taylor, Mathias Unberath, Ming-Yu Liu, and Chen-Hsuan Lin.
\newblock Neuralangelo: High-fidelity neural surface reconstruction.
\newblock In {\em Proceedings of the IEEE/CVF Conference on Computer Vision and Pattern Recognition}, pages 8456--8465, 2023.

\bibitem{li2023dynibar}
Zhengqi Li, Qianqian Wang, Forrester Cole, Richard Tucker, and Noah Snavely.
\newblock Dynibar: Neural dynamic image-based rendering.
\newblock In {\em Proceedings of the IEEE/CVF Conference on Computer Vision and Pattern Recognition}, pages 4273--4284, 2023.

\bibitem{liba2019handheld}
Orly Liba, Kiran Murthy, Yun-Ta Tsai, Tim Brooks, Tianfan Xue, Nikhil Karnad, Qiurui He, Jonathan~T Barron, Dillon Sharlet, Ryan Geiss, et~al.
\newblock Handheld mobile photography in very low light.
\newblock {\em ACM Trans. Graph.}, 38(6):164--1, 2019.

\bibitem{lipson2021raft}
Lahav Lipson, Zachary Teed, and Jia Deng.
\newblock Raft-stereo: Multilevel recurrent field transforms for stereo matching.
\newblock {\em arXiv preprint arXiv:2109.07547}, 2021.

\bibitem{liu2022semantic}
Yunfei Liu, Yu Li, Shaodi You, and Feng Lu.
\newblock Semantic guided single image reflection removal, 2022.

\bibitem{liu2020learning}
Yu-Lun Liu, Wei-Sheng Lai, Ming-Hsuan Yang, Yung-Yu Chuang, and Jia-Bin Huang.
\newblock Learning to see through obstructions.
\newblock In {\em IEEE Conference on Computer Vision and Pattern Recognition}, 2020.

\bibitem{lu2021omnimatte}
Erika Lu, Forrester Cole, Tali Dekel, Andrew Zisserman, William~T Freeman, and Michael Rubinstein.
\newblock Omnimatte: Associating objects and their effects in video.
\newblock In {\em CVPR}, 2021.

\bibitem{mertens2009exposure}
Tom Mertens, Jan Kautz, and Frank Van~Reeth.
\newblock Exposure fusion: A simple and practical alternative to high dynamic range photography.
\newblock In {\em Computer graphics forum}, volume~28, pages 161--171. Wiley Online Library, 2009.

\bibitem{mildenhall2018burst}
Ben Mildenhall, Jonathan~T Barron, Jiawen Chen, Dillon Sharlet, Ren Ng, and Robert Carroll.
\newblock Burst denoising with kernel prediction networks.
\newblock In {\em Proceedings of the IEEE conference on computer vision and pattern recognition}, pages 2502--2510, 2018.

\bibitem{mildenhall2022nerf}
Ben Mildenhall, Peter Hedman, Ricardo Martin-Brualla, Pratul~P Srinivasan, and Jonathan~T Barron.
\newblock Nerf in the dark: High dynamic range view synthesis from noisy raw images.
\newblock In {\em Proceedings of the IEEE/CVF Conference on Computer Vision and Pattern Recognition}, pages 16190--16199, 2022.

\bibitem{mildenhall2020nerf}
Ben Mildenhall, Pratul~P Srinivasan, Matthew Tancik, Jonathan~T Barron, Ravi Ramamoorthi, and Ren Ng.
\newblock Nerf: Representing scenes as neural radiance fields for view synthesis.
\newblock In {\em European conference on computer vision}, pages 405--421. Springer, 2020.

\bibitem{muller2022instant}
Thomas M{\"u}ller, Alex Evans, Christoph Schied, and Alexander Keller.
\newblock Instant neural graphics primitives with a multiresolution hash encoding.
\newblock {\em arXiv preprint arXiv:2201.05989}, 2022.

\bibitem{muller2021real}
Thomas M{\"u}ller, Fabrice Rousselle, Jan Nov{\'a}k, and Alexander Keller.
\newblock Real-time neural radiance caching for path tracing.
\newblock {\em arXiv preprint arXiv:2106.12372}, 2021.

\bibitem{nam2022neural}
Seonghyeon Nam, Marcus~A. Brubaker, and Michael~S. Brown.
\newblock Neural image representations for multi-image fusion and layer separation, 2022.

\bibitem{niklaus2020learned}
Simon Niklaus, Xuaner~Cecilia Zhang, Jonathan~T. Barron, Neal Wadhwa, Rahul Garg, Feng Liu, and Tianfan Xue.
\newblock Learned dual-view reflection removal, 2020.

\bibitem{park2011image}
Minwoo Park, Kyle Brocklehurst, Robert~T Collins, and Yanxi Liu.
\newblock Image de-fencing revisited.
\newblock In {\em Computer Vision--ACCV 2010: 10th Asian Conference on Computer Vision, Queenstown, New Zealand, November 8-12, 2010, Revised Selected Papers, Part IV 10}, pages 422--434. Springer, 2011.

\bibitem{shen2023light}
Zeqi Shen, Shuo Zhang, and Youfang Lin.
\newblock Light field reflection and background separation network based on adaptive focus selection.
\newblock {\em IEEE Transactions on Computational Imaging}, 9:435--447, 2023.

\bibitem{shih2015ghosting}
YiChang Shih, Dilip Krishnan, Fredo Durand, and William~T. Freeman.
\newblock Reflection removal using ghosting cues.
\newblock In {\em Proceedings of the IEEE Conference on Computer Vision and Pattern Recognition (CVPR)}, June 2015.

\bibitem{sitzmann2020implicit}
Vincent Sitzmann, Julien Martel, Alexander Bergman, David Lindell, and Gordon Wetzstein.
\newblock Implicit neural representations with periodic activation functions.
\newblock {\em Advances in Neural Information Processing Systems}, 33, 2020.

\bibitem{sun2021coil}
Yu Sun, Jiaming Liu, Mingyang Xie, Brendt Wohlberg, and Ulugbek~S Kamilov.
\newblock Coil: Coordinate-based internal learning for tomographic imaging.
\newblock {\em IEEE Transactions on Computational Imaging}, 7:1400--1412, 2021.

\bibitem{suvorov2021resolution}
Roman Suvorov, Elizaveta Logacheva, Anton Mashikhin, Anastasia Remizova, Arsenii Ashukha, Aleksei Silvestrov, Naejin Kong, Harshith Goka, Kiwoong Park, and Victor Lempitsky.
\newblock Resolution-robust large mask inpainting with fourier convolutions.
\newblock {\em arXiv preprint arXiv:2109.07161}, 2021.

\bibitem{tan2017joint}
Hanlin Tan, Xiangrong Zeng, Shiming Lai, Yu Liu, and Maojun Zhang.
\newblock Joint demosaicing and denoising of noisy bayer images with admm.
\newblock In {\em 2017 IEEE International Conference on Image Processing (ICIP)}, pages 2951--2955. IEEE, 2017.

\bibitem{tancik2020fourier}
Matthew Tancik, Pratul Srinivasan, Ben Mildenhall, Sara Fridovich-Keil, Nithin Raghavan, Utkarsh Singhal, Ravi Ramamoorthi, Jonathan Barron, and Ren Ng.
\newblock Fourier features let networks learn high frequency functions in low dimensional domains.
\newblock {\em Advances in Neural Information Processing Systems}, 33:7537--7547, 2020.

\bibitem{teed2021raft}
Zachary Teed and Jia Deng.
\newblock Raft-3d: Scene flow using rigid-motion embeddings.
\newblock In {\em Proceedings of the IEEE/CVF conference on computer vision and pattern recognition}, pages 8375--8384, 2021.

\bibitem{vogel2013piecewise}
Christoph Vogel, Konrad Schindler, and Stefan Roth.
\newblock Piecewise rigid scene flow.
\newblock In {\em Proceedings of the IEEE International Conference on Computer Vision}, pages 1377--1384, 2013.

\bibitem{wei2019single}
Kaixuan Wei, Jiaolong Yang, Ying Fu, David Wipf, and Hua Huang.
\newblock Single image reflection removal exploiting misaligned training data and network enhancements.
\newblock In {\em Proceedings of the IEEE/CVF Conference on Computer Vision and Pattern Recognition}, pages 8178--8187, 2019.

\bibitem{wronski2019handheld}
Bartlomiej Wronski, Ignacio Garcia-Dorado, Manfred Ernst, Damien Kelly, Michael Krainin, Chia-Kai Liang, Marc Levoy, and Peyman Milanfar.
\newblock Handheld multi-frame super-resolution.
\newblock {\em ACM Transactions on Graphics (TOG)}, 38(4):1--18, 2019.

\bibitem{xiong2019foreground}
Wei Xiong, Jiahui Yu, Zhe Lin, Jimei Yang, Xin Lu, Connelly Barnes, and Jiebo Luo.
\newblock Foreground-aware image inpainting.
\newblock In {\em Proceedings of the IEEE/CVF conference on computer vision and pattern recognition}, pages 5840--5848, 2019.

\bibitem{xue2015computational}
Tianfan Xue, Michael Rubinstein, Ce Liu, and William~T Freeman.
\newblock A computational approach for obstruction-free photography.
\newblock {\em ACM Transactions on Graphics (TOG)}, 34(4):1--11, 2015.

\bibitem{yang2022polynomial}
Guandao Yang, Sagie Benaim, Varun Jampani, Kyle Genova, Jonathan Barron, Thomas Funkhouser, Bharath Hariharan, and Serge Belongie.
\newblock Polynomial neural fields for subband decomposition and manipulation.
\newblock {\em Advances in Neural Information Processing Systems}, 35:4401--4415, 2022.

\bibitem{ye2022deformable}
Vickie Ye, Zhengqi Li, Richard Tucker, Angjoo Kanazawa, and Noah Snavely.
\newblock Deformable sprites for unsupervised video decomposition.
\newblock In {\em Proceedings of the IEEE/CVF Conference on Computer Vision and Pattern Recognition}, pages 2657--2666, 2022.

\bibitem{yu2021plenoctrees}
Alex Yu, Ruilong Li, Matthew Tancik, Hao Li, Ren Ng, and Angjoo Kanazawa.
\newblock Plenoctrees for real-time rendering of neural radiance fields.
\newblock In {\em Proceedings of the IEEE/CVF International Conference on Computer Vision}, pages 5752--5761, 2021.

\bibitem{yu20143d}
Fisher Yu and David Gallup.
\newblock 3d reconstruction from accidental motion.
\newblock In {\em Proceedings of the IEEE Conference on Computer Vision and Pattern Recognition}, pages 3986--3993, 2014.

\bibitem{zhu2023occlusion}
Chengxuan Zhu, Renjie Wan, Yunkai Tang, and Boxin Shi.
\newblock Occlusion-free scene recovery via neural radiance fields.
\newblock 2023.

\end{thebibliography}
